\documentclass[mnsc,nonblindrev]{informs3} 

\OneAndAHalfSpacedXI 

\usepackage{float}
\usepackage{subfig,graphicx}
\usepackage{algorithm}
\usepackage{algpseudocode}
\usepackage{amsmath}
\usepackage{bm}
\usepackage{hhline}

\usepackage{natbib}
 \bibpunct[, ]{(}{)}{,}{a}{}{,}%
 \def\bibfont{\small}%
\TheoremsNumberedThrough     
\ECRepeatTheorems

\EquationsNumberedThrough    

\MANUSCRIPTNO{MS-0000-0000.00}

\renewcommand{\theARTICLETOP}{} 
\begin{document}


\RUNAUTHOR{Xu, Lee, and Tan}


\RUNTITLE{Algorithmic Collusion or Competition}

\TITLE{Algorithmic Collusion or Competition: the Role of Platforms' Recommender Systems}

\ARTICLEAUTHORS{%
\AUTHOR{Xingchen (Cedric) Xu}
\AFF{Michael G. Foster School of Business, University of Washington, Seattle, WA 98195, \EMAIL{xcxu21@uw.edu}} 
\AUTHOR{Stephanie Lee}
\AFF{Michael G. Foster School of Business, University of Washington, Seattle, WA 98195, \EMAIL{stelee@uw.edu}} 
\AUTHOR{Yong Tan}
\AFF{Michael G. Foster School of Business, University of Washington, Seattle, WA 98195, \EMAIL{ytan@uw.edu}}
} 
\ABSTRACT{%
Recent scholarly work has extensively examined the phenomenon of algorithmic collusion driven by AI-enabled pricing algorithms. However, online platforms commonly deploy recommender systems that influence how consumers discover and purchase products, thereby shaping the reward structures faced by pricing algorithms and ultimately affecting competition dynamics and equilibrium outcomes. To address this gap in the literature and elucidate the role of recommender systems, we propose a novel repeated game framework that integrates several key components. We first develop a structural search model to characterize consumers’ decision-making processes in response to varying recommendation sets. This model incorporates both observable and unobservable heterogeneity in utility and search cost functions, and is estimated using real-world data. Building on the resulting consumer model, we formulate personalized recommendation algorithms designed to maximize either platform revenue or consumer utility. We further introduce pricing algorithms for sellers and integrate all these elements to facilitate comprehensive numerical experiments. Our experimental findings reveal that a revenue-maximizing recommender system intensifies algorithmic collusion, whereas a utility-maximizing recommender system encourages more competitive pricing behavior among sellers. Intriguingly, and contrary to conventional insights from the industrial organization and choice modeling literature, increasing the size of recommendation sets under a utility-maximizing regime does not consistently enhance consumer utility. This counterintuitive ``more is less" effect arises because expanding the recommendation set reduces the recommender system’s ability to effectively influence the pricing strategies of AI-driven algorithms through dynamic rewards. Moreover, the degree of horizontal differentiation moderates this phenomenon in unexpected ways. The ``more is less" effect does not arise at low levels of differentiation, but becomes increasingly pronounced as horizontal differentiation increases. This is because, in such contexts, consumers’ decisions are driven more by their inherent preferences than by price considerations. As a result, when the recommendation set is large, higher-priced products experience relatively smaller declines in their rewards and can sustain elevated prices, supported by the persistent demand engendered by horizontal preferences. These findings provide valuable insights for regulators, platform designers, and market participants, and they open up a range of promising avenues for future research.
}


\maketitle

\newpage
\section{Introduction}
\label{sec: intro}

\subsection{Motivation and Research Focus}
With the advancement of technology and the proliferation of sophisticated machine learning algorithms, the use of automated pricing systems has increased significantly across various sectors. For instance, a substantial share (34\%) of residential units in the United States currently utilize RealPage software, which provides instantaneous recommendations for optimal rental rates \citep{calder2023coordinated}. Artificial Intelligence (AI)-driven pricing programs, particularly those employing reinforcement learning techniques, have thus emerged as compelling alternatives to conventional pricing models \citep{spann2024algorithmic}. These AI-powered systems typically require minimal human intervention and exhibit remarkable adaptability to rapidly changing economic conditions.

However, both researchers and policymakers have voiced concerns about the potential detrimental effects of these algorithms on consumers, particularly regarding the phenomenon of algorithmic collusion. Algorithmic collusion refers to the capacity of algorithms to establish supra-competitive prices through repeated interactions \citep{calvano2020protecting}. In traditional markets without AI-based pricing, game theory suggests that supra-competitive prices occur when businesses collude to set artificially elevated prices \citep{athey2004collusion}. Since each seller is incentivized to undercut these agreed-upon prices, the sustainability of a supra-competitive price depends on (1) the cartel’s ability to effectively punish deviators and (2) each firm’s perception of other members’ reactions. In the context of algorithmic collusion, however, these mechanisms are no longer required: sellers merely set a revenue-optimizing objective, and the pricing algorithms can independently learn to charge supra-competitive prices without direct communication or explicit instructions on how to collude. In such scenarios, these prices often remain stable, with algorithms consistently maintaining higher rates at the expense of consumers \citep{calvano2020protecting}.

Given the increasing attention to this issue, related legislative bills and legal cases have begun to emerge. For example, the ``Preventing Algorithmic Collusion Act'' was proposed in 2024\footnote{See \url{https://www.congress.gov/bill/118th-congress/senate-bill/3686} for more details.}, and RealPage has been sued for employing pricing algorithms that potentially harmed millions of renters\footnote{See \url{https://www.cbsnews.com/news/realpage-lawsuit-price-fixing/} for more details.}. In parallel, a growing body of literature has attempted to address this area. On the empirical front, research has provided suggestive evidence that the adoption of pricing algorithms correlates with increased price synchronization among firms and elevated prices \citep{calder2023coordinated, wieting2021algorithms, assad2020algorithmic}. In order to better understand the underlying mechanisms, numerical experimental approaches have been employed since the seminal work of \cite{calvano2020artificial} \citep{dou2024ai, rocher2023adversarial}.

In online marketplaces such as Amazon, pricing algorithms are also widely utilized \cite{chen2016empirical}. Nevertheless, despite existing scholarly efforts and regulatory interventions, the role of recommender systems—which are extensively used in online marketplaces—remains largely overlooked in the context of algorithmic pricing games. Conceptually, recommender systems can alter product exposure and thus influence the rewards obtained by pricing algorithms, ultimately shaping pricing dynamics and welfare distributions. Recognizing this gap, we aim to address it by examining \textit{how various recommendation algorithms affect pricing dynamics and equilibrium outcomes in the algorithmic pricing game.}

\subsection{Research Framework}

To examine the algorithmic pricing game under the influence of recommender systems, it is essential to develop a modified repeated game framework. \citet{calvano2020artificial} provides a seminal foundation for the study of algorithmic collusion, offering a framework that has been widely adopted and adapted in subsequent research \citep{dou2024ai, rocher2023adversarial, banchio2022artificial}. \cite{calvano2020artificial} specifies three key elements: (1) the consumer demand model, where consumer decisions are governed by a canonical logit model; (2) the pricing algorithms, in which sellers employ basic Q-learning algorithms to adaptively set prices; and (3) the game sequence, where at each discrete period, sellers simultaneously make pricing decisions and observe the resulting demand responses.

Given our research focus, we need to specify and integrate the recommender system as a fourth element within the framework. Furthermore, it is also necessary to adjust the other three components accordingly to ensure coherent integration, which presents certain challenges. In the presence of recommender systems, a key challenge is to model consumer decision-making when consumers are exposed to multiple recommended products simultaneously. Typically, consumers observe a ranked list of top-$K$ products and need to expend time and effort to evaluate the options of interest \citep{moraga2023consumer, mortensen2011markets}. Additionally, position biases may influence their choices \citep{guo2019pal, ursu2018power}. To accurately capture these considerations and establish a solid economic foundation, we develop a customized structural search model that incorporates heterogeneous utility functions and search costs \citep{chung2024simulated, ursu2023sequential}. We then estimate the model parameters using real-world data, employing an innovative combination of Hermite approximation methods and simulated maximum likelihood estimation to reduce computational complexity. The calibrated model subsequently generates demand responses within the repeated game framework.

With this demand model in place, we also specify a recommender system that optimizes the platform’s objective, such as overall revenue or consumer utility \citep{compiani2024online, wang2024recommending, zhang2024algorithmic, li2024deep}. However, when the demand model is grounded in a sophisticated framework such as sequential search, closed-form expressions for consumer choice probabilities and optimal recommendations are generally unavailable. As a result, we follow established optimization literature and employ numerical integration methods to identify recommendation actions that maximize the specified objective function \citep{compiani2024online}.

For the pricing algorithm and game sequence, we adopt the Q-learning approach and the simultaneous game structure outlined by \cite{calvano2020artificial}. Nonetheless, Q-learning inherently assumes forward-looking behavior, prompting us to examine an alternative in the form of a simpler bandit algorithm to ensure robustness. Similarly, to further test the flexibility of our framework, we also consider an asynchronous game sequence as an additional robustness check.

By integrating all of these elements, we can analyze the impact of recommender systems on algorithmic pricing. However, even the original framework proposed by \cite{calvano2020artificial} poses significant analytical challenges. In a multi-agent environment, each agent’s setting is non-stationary, rendering both the learning of strategies and the determination of equilibria analytically intractable. Introducing a recommender system, alongside adjustments to the other model components, further complicates the analysis. In particular, the absence of closed-form expressions for consumer choice probabilities precludes straightforward derivation of sellers’ demands and revenues given any set of prices and recommendations.

To address these complexities, we extend the approach of \cite{calvano2020artificial}, relying on numerical simulations to trace the price evolution and identify equilibrium outcomes. In each period, after sellers have set prices and the platform has determined recommendations, we use the empirical distribution of consumer attributes and the calibrated consumer demand model to compute each seller’s expected reward. This involves multiple layers of numerical integration. Through this procedure, we can efficiently execute simulations and conduct counterfactual analyses by altering any component within the proposed framework.

\subsection{Findings}

Employing the aforementioned setups, we adjust various parameters within the framework to evaluate how such changes influence price dynamics and equilibrium outcomes. Turning to our primary focus—the recommender system—from an economic perspective, it optimizes the platform’s objective by selecting and potentially ranking a subset of products to display to consumers from the full set of available options \citep{feldman2022customer}.

Previous research on product display algorithms in two-sided markets predominantly considers two key objectives\footnote{It is noteworthy that the technical literature on recommender systems primarily categorizes these systems based on their underlying methodologies, such as classical collaborative filtering and deep learning techniques \citep{zhang2019deep, adomavicius2005toward}. Objectives (outcomes) are generally self-explanatory and thus receive limited attention within this stream of literature. In contrast, our research emphasizes these objectives from an economic perspective.}. The first, \textit{Revenue-Maximization}, seeks to maximize total platform-wide revenue, thereby aligning with the platform’s economic incentives through commission fees. The second, \textit{Utility-Maximization}, aims to enhance consumer utility, which may indirectly benefit the platform by improving consumer retention \citep{compiani2024online, feldman2022customer, derakhshan2022product, ursu2018power}. Since different objectives can substantially alter product exposure and consequently influence sellers’ learned strategies as well as price dynamics, we pose the following research question:

\begin{quote}
\textit{RQ1: How does the objective of the recommender system affect the algorithmic pricing game?}
\end{quote}

We address this research question through numerical simulations based on our proposed framework, which comprises the following components: a structural search model calibrated with empirical data to generate demand responses; Q-learning algorithms to determine pricing decisions; recommender systems designed to generate recommendation sets for each consumer in alignment with the platform's objectives; and a simultaneous game sequence that integrates the behaviors of all participants within a repeated game context. 

We conduct numerical simulations under two distinct recommender system objectives—revenue maximization and utility maximization—and compare these results to a baseline scenario in which product display is randomized. Our findings reveal that the objectives of recommender systems significantly influence both pricing trajectories and the resulting price equilibrium. Specifically, revenue-maximization recommender systems tend to exacerbate algorithmic collusion, leading to elevated market prices. In contrast, utility-maximization systems yield non-collusive outcomes characterized by lower equilibrium prices. Furthermore, we explore the impact of varying the relative weights assigned to revenue and utility objectives on market outcomes. The results demonstrate a monotonic relationship between these weights and the resulting prices, further reinforcing the crucial role of recommender system objectives in shaping algorithmic pricing dynamics.

Beyond the objective of the recommender system, another significant factor it can predefine is the number of products displayed. This dimension holds considerable economic importance, as it directly influences search friction and thus the intensity of competition among sellers \citep{moraga2023consumer, hong2006using}. Guided by the antitrust and algorithmic collusion literature’s emphasis on enhancing consumer welfare \citep{baker2019antitrust}, one may naturally question whether displaying more products can effectively achieve this goal. The underlying rationale is that, with a wider range of product options, consumers can bypass items they deem less desirable and choose those that offer higher utility. This reasoning is supported by the choice modeling literature, which suggests that the welfare derived from a product set increases as more products are made available \citep{train2009discrete}. To investigate whether this logic holds true when pricing algorithms are in play, we pose our second research question:

\begin{quote}
\textit{RQ2: Does displaying more products when using the utility-maximization recommender system always benefit consumers?}
\end{quote}

By running simulations with varying numbers of displayed products, we uncover a counterintuitive finding: under a utility-maximization objective, increasing the number of displayed products can sometimes reduce consumer utility. This outcome emerges from two competing mechanisms. On one hand, offering more products improves consumers’ chances of discovering those that yield higher utility. On the other hand, expanding the set diminishes the platform’s leverage in influencing prices through recommendations. Consequently, high-priced products retain exposure and continue to attract some consumers, experiencing relatively smaller reductions in their rewards and thus sustaining higher equilibrium prices. When this latter effect dominates, providing more products ultimately harms consumer welfare.

Building on the previous rationale, horizontal differentiation emerges as another key factor influencing the effects of increasing the number of displayed products. When products differ only minimally, consumers tend to select the lowest-priced option, rendering additional offerings largely inconsequential. On the contrary, greater horizontal differentiation may allow consumers to find products that better match their preferences, thus potentially enhancing their utility through a broader set of options. Given the importance of horizontal differentiation in price competition and industrial organization research \citep{colombo2013product, thomadsen2007costly, tyagi1999relationship}, we ask our third research question:

\begin{quote}
\textit{RQ3: How does horizontal differentiation moderate the relationship between the number of displayed products and consumer utility?}
\end{quote}

By adjusting the horizontal differentiation parameter and examining the impact of increasing the number of displayed products, we uncover another counterintuitive moderation effect. When horizontal differentiation is low, offering more product options benefits consumers by enabling them to find preferred items. However, as horizontal differentiation increases, the ``more is less'' effect emerges. While greater differentiation enhances the potential gains from searching among a broader range of products to identify those closely aligned with consumer preferences, it also enables higher-priced offerings to retain demand and maintain elevated prices. Consequently, at high levels of horizontal differentiation, both mechanisms intensify, and the disadvantage of sustaining higher-priced products may outweigh the potential utility gains derived from increased product variety. Therefore, contrary to conventional intuition, providing more products may diminish consumer utility under high horizontal differentiation, while enhancing it when horizontal differentiation is low.

\subsection{Summary of Contributions}

Our work contributes methodologically, theoretically, and practically to the literature, with subsequent sections providing further details on each dimension.

From a \textit{methodological} perspective, we incorporate recommender systems into a repeated game framework and appropriately adjust other model components. In doing so, we address the estimation of a complex structural search model by introducing the Hermite method into the simulated maximum likelihood procedure. We also derive the sequence of the game and equilibrium using multi-layer numerical integration, while relaxing several key assumptions from \cite{calvano2020artificial}. The resulting framework offers a valuable foundation for future analyses of analogous research questions.

From a \textit{theoretical} perspective, we extend the algorithmic collusion literature by highlighting the critical role of recommender systems, which alter product exposure and thus influence the rewards that pricing algorithms receive. This interplay ultimately affects pricing dynamics and welfare outcomes. Additionally, we enrich the economics of recommender systems literature by incorporating adaptive, AI-based pricing rather than presupposing static or fully informed seller-driven pricing strategies. In other words, we demonstrate that recommender systems can reshape economic outcomes not only by shifting consumer behavior but also by altering the dynamic incentives that drive pricing algorithms, thus affecting welfare distributions. Furthermore, this study contributes to the broader economics of AI literature by taking a first step toward examining the aggregate economic effects of cross-category AI-AI interactions—specifically, those between recommender systems and multiple pricing algorithms—rather than treating each AI intervention in isolation.

From a \textit{practical} perspective, our results yield important insights for regulators, platforms, and sellers. For instance, when regulators attempt to mitigate supracompetitive pricing through antitrust measures, focusing exclusively on the providers of pricing algorithms may prove insufficient. Platforms, such as Amazon, may need to refine their recommendation policies by jointly considering consumer responses and the strategic behavior of pricing algorithms. Similarly, sellers might benefit from coordinating product design and pricing algorithm strategies to account for the platform’s evolving recommendation criteria.

In the subsequent sections, we first review the related literature and identify our \textit{theoretical contributions} to each research stream in Section \ref{sec: literature}. In Section \ref{sec: base_setup}, we present our repeated game framework, detailing its components, the challenges addressed, and our \textit{contributions to frameworks and models}. Section \ref{sec: estimation} outlines our parameterization strategy, describes the empirical data, and explains the estimation procedures employed to calibrate the structural search model. Building on this foundation, we offer our findings through numerical simulations in Section \ref{sec: results}, followed by a series of extended analyses in Section \ref{sec: Extended} to test the robustness of these results. Finally, Section \ref{sec: conclusion} concludes by discussing the \textit{practical implications} of our study and suggesting directions for \textit{future research}.

\section{Related Literature}
\label{sec: literature}
By incorporating recommender systems into the repeated pricing games among sellers that utilize AI-driven pricing algorithms, our study contributes to the literature on the economics of AI, algorithmic collusion, and the economics of recommender systems. In each subsection, we first review the relevant literature and subsequently highlight our specific contributions to each research stream.

\subsection{Economics of AI}

Broadly speaking, our study is situated within the domain of the economics of AI, wherein economic principles are applied to evaluate, design, and implement AI techniques and architectures \citep{agrawal2019economics}. A growing body of literature examines the impact of various AI systems across diverse contexts, such as AI assistants for customer response \citep{brynjolfsson2023generative}, matching algorithms in the gig economy \citep{liu2024unintended}, AI copilots in software development \citep{peng2023impact}, and AI applications in creative works \citep{boussioux2024crowdless, gao2024quantifying}. These insights inform the design of future AI systems to be more responsible and ethical, thereby enhancing overall welfare and promoting its equitable distribution \citep{maslej2023artificial}.

The focus of this work, AI-based pricing algorithms (particularly algorithmic collusion) and recommender systems, represents two significant subfields within the economics of AI literature. However, these areas are typically studied separately, resulting in distinct streams of research as discussed in the subsequent two subsections \citep{calvano2020artificial, fleder2009blockbuster}. By jointly examining the economic impacts of recommender systems and pricing algorithms, we contribute to the broader literature by investigating a unique phenomenon concerning the economics of AI-AI interactions.

\subsection{Algorithmic Collusion and Antitrust Regulation}

More specifically, this study is related to and contributes to burgeoning research on algorithmic collusion. With the prevalence of learning algorithms, scholars have investigated the impact of AI-based pricing algorithms on market competition and price equilibrium, particularly since the seminal work by \citet{calvano2020artificial}. Antitrust policymakers have also expressed concerns regarding algorithmic collusion, which could consequently lead to adverse effects on consumer welfare \citep{robertson2022antitrust, colangelo2021artificial, fzrachi2019sustainable, schwalbe2018algorithms}.

To describe and understand algorithmic collusion, some early works have employed theoretical models to illustrate how simple pricing strategies can decipher competitors' behavior and lead to collusive outcomes \citep{miklos2019collusion, salcedo2015pricing}. However, as algorithms become more sophisticated, these models become analytically intractable, thereby necessitating numerical simulations within the repeated game framework \citep{calvano2020artificial}. Subsequently, several studies have explored algorithmic collusion in various market contexts, such as auctions \citep{banchio2022artificial} and securities \citep{dou2024ai, colliard2022algorithmic}. Additionally, certain empirical evidence has been documented, supporting instances of correlated and elevated prices among multiple vendors, although the underlying mechanisms are challenging to empirically verify \citep{calder2023coordinated, wieting2021algorithms, assad2020algorithmic}.

Although research on algorithmic collusion is active, it is noteworthy that existing studies on collusion in algorithmic pricing have not thoroughly considered the influence of platform recommendation systems, which directly impact product exposure and profitability in online marketplaces. This study contributes to the algorithmic collusion literature by demonstrating that the platform's recommendation algorithm plays a pivotal role in shaping the competitive landscape of the market and influencing the degree of collusion among sellers. Furthermore, the study offers an alternative explanation for the observed correlated prices in empirical studies, suggesting that the platform's recommendation mechanism may drive this correlation. Consequently, for antitrust regulations targeting supracompetitive prices in online markets with recommender systems, it may be more equitable to attribute responsibilities to both platforms and sellers.

\subsection{Economics of Recommender Systems}
\label{subsec: lit_recsys}

Our study is closely aligned with the literature on the economics of recommender systems, which investigates the downstream impacts of these systems on various economically significant outcomes, driven by the complexity of consumer behavior in real-world settings.

For instance, \citet{fleder2009blockbuster} examine how recommender systems influence both individual-level and market-level sales diversity through analytical models and simulations. Building on this, \citet{lee2019recommender} empirically test the impact of recommender systems on diversity using field experiments. More recently, \citet{lee2021product} documents the heterogeneity in the economic impact of recommender systems, while \citet{li2022recommender} explore the causal pathways underlying these effects. In addition to studies that focus on the economic benefits that recommendations provide to sellers and platforms, there is also research that considers consumer welfare \citep{wan2024product, zhang2021welfare} and examines the dynamic interactions between consumer choices and recommender systems \citep{zhou2023longitudinal}. 

However, a significant gap in the existing literature is that previous studies on recommender systems have typically treated product pricing as an exogenous variable. Only a few exceptions have analytically explored the effects of recommender systems on pricing decisions and market equilibrium. In these analyses, sellers are generally modeled as revenue-seeking agents, while the platform may possess different incentives, such as revenue maximization or utility optimization \citep{zhou2023competing, li2020informative, li2018recommender}. Nonetheless, these studies are predominantly static in nature and assume that sellers possess complete market information \citep{zhou2023competing, li2020informative, li2018recommender}.

In this context, our study makes a novel contribution by allowing prices to be dynamically determined by AI-driven pricing algorithms, which learn to set prices through interaction with the market. We then investigate the economic impact of recommender systems on price dynamics, sellers' benefits, and consumer utility from a holistic perspective. This approach bridges the gap in the literature by integrating the behavior of pricing algorithms with the economics of recommender systems, thereby providing a more comprehensive understanding of their interplay in modern online platforms.

\section{Dynamic Pricing Competition Moderated by the Platform’s Recommendation System}
\label{sec: base_setup}

\subsection{General Repeated Game Framework}
\label{subsec: framework}

Despite its critical importance, the role of platform recommendations in shaping algorithmic pricing strategies remains underexplored in the existing literature. Previous studies on algorithmic collusion, largely following the framework established by \cite{calvano2020artificial}, typically include a logit demand model, employ Q-learning as the pricing algorithm, and consider a simultaneous game sequence. To more effectively address our research questions, we develop a novel repeated game framework that incorporates the platform’s recommendation system and modifies these three elements accordingly.

This section presents the overall structure of our repeated game model, specifies the sequence of play, and provides a concise summary of the principal challenges encountered. In the subsections that follow, we discuss each key element in detail: Section \ref{subsec: Economic} introduces the demand model, Section \ref{subsec: Q-learning} details the pricing algorithm, and Section \ref{subsec: Recsys_main} outlines the platform’s recommendation algorithm. For ease of reference, Table \ref{tab:notations} compiles the central notations employed in this study. Although additional symbols may emerge as we formally characterize and estimate the customer demand model, we refrain from listing them here to preserve clarity.

\begin{table}[ht] \centering 
\TABLE{Summary of Notations for the Main Model\label{tab:notations}}
{
\begin{tabular}{ p{2cm} p{12cm}}
\\[-1.8ex]\hline  
\hline \\[-1.8ex]  
Notation & Explanation \\
\hline \\[-1.8ex] 
$t$ & Time period $t$\\
$J$ & The number of products available on the market\\
$K$ & The number of products shown to each consumer\\
$j$ & A product $j$ (with $j = 0$ representing the outside option)\\
$p_j^t$ & The price of product $j$ at period $t$\\
$\bm{X_j}$ & The other attributes of product $j$ which are stable across periods \\
$D_j^t$ & Demand of product $j$ at period $t$\\
$\pi_j^t$ & Revenue of product $j$ at period $t$\\
$i$ & A Consumer $i$ (the index is reordered every period to simplify the notation)\\ 
$\bm{A_i}$ & The consumer $i$'s attributes\\
$\bm{B_i}$ & A set of products shown to the consumer $i$\\
$\beta_i$ & The consumer's price sensitivity \\
$\bm{\Gamma}$ & Other parameters linking product' attributes to utility\\
$\bm{\Theta}$ & Other preference parameters linking consumer's attributes to utility\\
$pos_{i,j}$ & Ranking (Position) of product $j$ seen by consumer $i$\\
$c_{i,j}$ & The search cost of product $j$ for consumer $i$\\
$\tau_{i}$ & The constant term of consumer $i$'s search cost\\
$\varphi$ & The ranking effect parameter of the search cost\\
$u_{i,j}$ & The consumption utility of product $j$ for consumer $i$\\
$v_{i,j}$ & The expected pre-search utility of product $j$ for consumer $i$\\
$\xi_{i,j}$ & The post-search utility shock of product $j$ for consumer $i$\\
$z_{i,j}$ & The reservation utility of product $j$ for consumer $i$\\
$\mu$ & The horizontal differentiation level on the market\\
$S_t$ & State of the system at the period $t$, which is the price vector at $t-1$\\
$Q_j$ & Product $j$'s pricing strategy (characterized by a Q matrix in Q-learning)\\
$TU^t$ & Total Consumer Utility at period $t$\\
$TD^t$ & Total Demand at period $t$\\
$\Pi^t$ & Total Revenue at period $t$\\
$r_j^t$ & Product $j$'s reward at period $t$\\
$m^t$ & Platform's reward at period $t$\\
$\alpha$ & Learning rate of the pricing algorithm\\
$\delta$ & Discount rate of the pricing algorithm\\
$\epsilon$ & Exploration rate of the pricing algorithm\\
$\omega$ & Decay rate of exploration rate\\
\\
\hline \\
\end{tabular}}
{}
\end{table}

Specifically, consider a marketplace comprising \( J \) heterogeneous products (\( j \in \{1, 2, \ldots, J\} \))\footnote{We assume a one-to-one mapping between sellers and products. If a seller offers multiple products with distinct dynamic pricing strategies, each product can be treated as a separate seller.}, each associated with an attribute vector \(\bm{X}_j\), as well as an outside option (\( j = 0 \)). At each discrete time period \( t \), sellers independently and simultaneously set their prices \( p_j^t \). Concurrently, a cohort of consumers, indexed by \( i \) and each characterized by an attribute vector \(\bm{A}_i\), enters the platform. The platform then constructs a recommendation set \(\bm{B}_i\) for each consumer and establishes product rankings \( pos_{i,j} \) to optimize its own objective. Upon viewing these recommendations, each consumer \( i \) undertakes a search process and ultimately purchases product \( k \), yielding utility \( u_{i,k} \). As a result, each seller realizes a reward \( r_j^t \), enabling it to iteratively refine its pricing algorithm over time. Meanwhile, the platform also accrues a reward \( m^t \). This cyclical process is summarized below in Algorithm \ref{alg1}.

\begin{algorithm}[h!]
\caption{General Repeated Game Process}\label{alg1}
\begin{algorithmic}[1]
\Require {Initial price $p_{1}^{0}$,..., $p_{J}^{0}$; product attribute $\bm{X_1}$,...,$\bm{X_J}$; initial pricing strategies $Q_1$,..., $Q_J$; consumer attribute $\bm{A_1^t}$,...,$\bm{A_N^t}$}; consumer decision model $CM$; initial recommendation strategy $RS$.

\While {$t\leq t_{max}$}
\State {\textbf{\textit{Step 1: Pricing Decision}}}
\For{each seller $j$:}
\State $p_j^t$ is made based on $Q_j$.
\EndFor
\State {\textbf{\textit{Step 2: Recommendation and Demand Realization}}}
\For{each consumer $i$:} 
\State Consumer $i$ sees a list of products with ranking $\bm{B_i}$ which is determined based on $p_{1}^{0}$,..., $p_{J}^{0}$, $\bm{X_1}$,...,$\bm{X_J}$, and $\bm{A_i^t}$ by $RS^t$.
\State Consumer $i$ makes the search and purchase decisions following the decision model $M$.
\EndFor
\For{each seller $j$:} 
\State Demand $D_j^t$ and Profits $R_j^t$ are realized.
\EndFor
\State Total utility $TU$, total demand $TD^t$, and total Revenue $\Pi^t$ on the platform are realized.
\State {\textbf{\textit{Step 3: Algorithm Update}}}
\For{each seller $j$:} 
\State Pricing strategy $Q_j$ is updated based on the seller's objective $r_j^t$.
\EndFor
\State Platform also realizes the reward $m^t$.
\State $t \leftarrow t+1$.
\EndWhile
\end{algorithmic}
\end{algorithm}

The repeated game process described above builds on the foundational work of \cite{calvano2020artificial}, a benchmark study in the field of algorithmic collusion that has influenced numerous subsequent investigations \citep{dou2024ai, rocher2023adversarial}. However, incorporating recommendation systems into this framework introduces a set of additional methodological complexities.

First, the introduction of recommender systems necessitates modeling consumer decision-making amid multiple simultaneously presented product recommendations. Consumers typically encounter a ranked list of products and need to devote time and effort to assess those of interest \citep{moraga2023consumer}. Additionally, position biases may influence their choices \citep{jeunen2021top, guo2019pal, ursu2018power, hofmann2014effects}. In response, we develop a model that incorporates both consumer search behaviors and position biases, as discussed in Section \ref{subsec: Economic}.

Second, consumers exhibit inherent heterogeneity, requiring sufficient flexibility in both the demand model and the associated personalized recommendation algorithms \citep{chung2024simulated, morozov2021estimation, zhang2019deep}. Accounting for such realities complicates the estimation of demand models and the optimization of recommendation strategies. We address these challenges in Sections \ref{subsec: Economic}, \ref{subsec: Recsys_main}, and \ref{sec: estimation}.

Finally, while the framework introduced by \cite{calvano2020artificial} is notably adaptable and involves relatively fewer stringent assumptions, it still presupposes forward-looking pricing algorithms and simultaneous decision-making among sellers. We critically examine these assumptions and relax them, thereby extending the model’s scope and robustness. The details of these modifications and their implications are presented in Section \ref{sec: Extended}.

\subsection{Consumer Demand Model}
\label{subsec: Economic}

In this subsection, we examine a structural search model to elucidate the decision-making processes of consumers when presented with platform-based recommendations. We adopt the sequential search framework for several reasons. First, consumers typically lack complete information about all available products in the market and face significant costs when seeking information \citep{moraga2023consumer, mortensen2011markets}. This aligns with the objectives of recommender systems, which aim to alleviate the challenges of information overload and search frictions \citep{zhang2019deep}. The sequential search model addresses these issues by explicitly incorporating the search costs incurred by consumers in accessing product information \citep{ursu2023sequential, weitzman1979optimal}. Second, when multiple products are recommended, position bias may emerge, a phenomenon that the sequential search framework can effectively capture \citep{chung2024simulated, ursu2018power}. Third, the sequential search model allows for the incorporation of heterogeneous preferences and search costs among consumers, thereby facilitating personalized recommendations within this framework \citep{ursu2023sequential, morozov2021estimation}.

Specifically, each consumer \(i\) is presented with a recommendation set \(\bm{B_i}\), resulting from the platform's top-\(K\) recommendation algorithm (\(K \le J\))\footnote{In this section, the time index \(t\) is omitted for simplicity, although \(p_j^t\) and \(A_i^t\) may vary over time.}. Here, \(\bm{B_i}\) denotes a consumer-specific subset of all products (\(j = 1, 2, \ldots, J\)) in the marketplace. The position of product \(j\) for consumer \(i\) is represented by \(pos_{i,j}\) (\(pos_{i,j} \in \{1, 2, \ldots, K\}\)). From the consumer's perspective, the value of each product is quantified by the purchase utility function. This utility, \(u_{i,j}\), derived by consumer \(i\) from product \(j\), is decomposed into two components:
\begin{equation}
u_{i,j} = v_{i,j} + \xi_{i,j}
\end{equation}

The first component, \(v_{i,j}\), represents the utility observable on the search results page \textit{prior} to clicking. In contrast, the second component, \(\xi_{i,j}\), which follows a normal distribution \(N(0, \delta_{\xi}^2)\), denotes the utility revealed only \textit{after} navigating to the product's webpage.

Moreover, each product \( j \) is characterized by its price \( p_j \) and a vector of attributes \( \bm{X_j} \), which are disclosed to consumers prior to their search. Similarly, consumer \( i \) has a series of attributes \( \bm{A}_i \). Consequently, the decision-making context for each consumer \( i \) includes the following elements: the consumer \( i \) with attributes \( \bm{A_i} \), product \( j \) with its price \( p_j \) and attributes \( \bm{X_j} \) (\( j \in \bm{B_i} \)), and the ranking of these product \( \psi_{i,j} \). Consistent with prior research on sequential search models, we postulate a linear parametric relationship between these observed attributes and the consumer’s pre-search expected utility \citep{chung2024simulated, ursu2018power}:
\begin{equation}
    v_{i,j} = \beta_i * p_j + \bm{\Gamma^{\prime}} * \bm{X_j} + \bm{\Theta^{\prime}} * \bm{A_i}
\end{equation}

In addition to the utility derived from purchasing products, consumers incur a cost associated with the search required to ascertain $\xi_{i,j}$. This cost is encapsulated by the search cost function $c_{i,j}$, which, in accordance with extant literature, adheres to an exponential function contingent upon the product's ranking:
\begin{equation}
c_{i,j} = \exp{(\tau_i + \varphi \cdot pos_{i,j})}
\end{equation}

Given consumers' uncertainty regarding the precise magnitude of $\xi_{i,j}$—with only its probabilistic distribution known—it becomes necessary for consumers to balance the potential utility gains from discovering products that offer higher purchase utility, $u_{i,j}$, against the associated search costs, $c_{i,j}$. This trade-off in consumer decision-making is encapsulated by the concept of reservation utility \citep{weitzman1979optimal}. The reservation utility for a product, denoted by $z_{i,j}$, is defined as the utility level at which the consumer is indifferent between continuing the search for product $j$ or obtaining a utility of $z_{i,j}$ with certainty. Mathematically, this represents the equivalence between the marginal utility derived from searching for product $j$ and the marginal cost incurred, expressed as:
\begin{equation}
\begin{aligned}
    c_{i,j} &= \int_{z_{i,j}}^{\infty}\left(u_{i,j}-z_{i,j}\right) f(u_{i,j}) d u_{i,j} \\
    &=  \int_{z_{i,j}}^{\infty}\left(v_{i,j} + \xi_{i,j} -z_{i,j}\right) f(\xi_{i,j}) d \xi_{i,j}
\end{aligned}
\end{equation}

Incorporating purchase utility and search costs, the framework that governs consumer search and selection behavior is delineated by three fundamental rules: the selection, stopping, and choice rules. These rules comprehensively characterize consumer behavior, as detailed below:

\begin{enumerate}
    \item \textbf{Selection Rule}: This rule stipulates that consumers rank products in descending order based on their reservation utilities, denoted by $z_{i,j}$. This ranking determines the sequence in which products are searched. Specifically, the product with the highest reservation utility, $z_{i,j}$, is selected as the next candidate for evaluation.

    \item \textbf{Stopping Rule}: After evaluating the set of products \( \bm{H_i} \) and observing their post-search utilities\footnote{Consistent with prior literature, we assume that consumers are aware of their outside option before searching recommended alternatives. Consequently, we further assume $H_i(0) = 0$ for notational simplicity \citep{ursu2023sequential}.}, consumer \(i\) needs to decide whether to continue the search or terminate it. The search proceeds if the highest utility realized from the searched products \(\bm{B_i}\) is less than the maximum reservation utility among the yet-to-be-searched options, \(\bm{B_i} \setminus \bm{H_i}\):
    \begin{equation}
        \max _{j \in \bm{H_i}} u_{i,j} < \max _{j \in \bm{B_i} \setminus \bm{H_i}} z_{i,j}
    \end{equation}

    Conversely, the search is terminated if the above condition is not met. In other words, the maximal utility among all the searched products exceeds the reservation utilities of all the unsearched products. Additionally, the search process also terminates if all options have been searched.
    
    \item \textbf{Choice Rule}: Upon cessation of the search, the consumer selects the product \(k\) that exhibits the highest utility from all the searched products:
    \begin{equation}
        u_{i,k} \geq \max _{j \in \bm{H_i}} u_{i,j}
    \end{equation}
\end{enumerate}

By integrating these rules, consumers make search and purchase decisions based on available information. The entire process is visualized in Appendix \ref{appendix:visualization}. Section \ref{sec: estimation} provides further details on how to estimate this model and derive parameters from the data.

\subsection{Pricing Algorithms Adopted by Sellers}
\label{subsec: Q-learning}

In modeling the pricing algorithm, we employ the canonical Q-learning for investigating pricing games, consistent with prior research on algorithmic collusion \citep{dou2024ai, wang2023algorithms, colliard2022algorithmic, banchio2022artificial}. The selection of the classical Q-learning approach is motivated by several compelling factors \citep{calvano2020artificial}. First, Q-learning is extensively utilized in practice and serves as the foundation for numerous state-of-the-art reinforcement learning algorithms \citep{li2023reinforcement}, including DQN \citep{mnih2015human}, Double DQN \citep{van2016deep}, Double-Dueling DQN \citep{wang2023deep}, and many advanced variants \citep{silver2018general, silver2016mastering}. Consequently, studying Q-learning provides insights into a broad spectrum of algorithmic pricing models that build upon its adaptations. Second, Q-learning models the reward structure non-parametrically and requires only a minimal number of parameters. Its simplicity, relative to neural network-based algorithms, requires fewer assumptions and facilitates a clearer interpretation of the parameters' impact on game dynamics. Third, Q-learning's capacity to operate without prior knowledge of the economic environment aligns with real-world scenarios where sellers possess limited market information, thereby enhancing the practical applicability of our pricing model.

We now provide a detailed description of our Q-learning algorithm and introduce the relevant notation for our framework. Building on the foundational work on Q-learning, we define the agent's state, action, immediate reward, and objective to determine its optimal strategy within the framework of Markov decision processes \citep{watkins1989learning}. In each discrete period $t$, the pricing agent $j$ (i.e., seller $j$ on the platform) observes the current state variable $s^t$ and selects a pricing action $p_j^t$ from a set of feasible actions $\bm{P}$. Although the pricing agent has access to historical pricing data, to streamline calculations and maintain consistency with prior studies, we assume a one-period memory \citep{calvano2020artificial}, resulting in the state being represented by the prices of all products in the previous period: $s^t = (p_1^{t-1}, \dots, p_J^{t-1})$. Subsequently, the pricing agent receives a reward $r_j^t$ from the environment and observes the complete set of prices transitioning to the next period, denoted by $s^{t+1} = (p_1^{t}, \dots, p_J^{t})$. In the context of our pricing game, the reward corresponds to the seller's revenue generated by the demand model. The primary objective of the pricing agent $j$ is to maximize the total discounted cumulative reward (i.e., revenue for seller $j$), expressed as $\mathbb{E} \left[ \sum_{t=0}^{\infty} \delta^t r_j^t \right]$, where $\delta$ represents the discount factor.

Since each agent's action influences not only the immediate reward but also the subsequent state transition, it is imperative for the agent to adopt a forward-looking perspective that accounts for the value of each state \( s \)\footnote{This assumption will be relaxed in Section \ref{sec: Extended}.}. This state-specific value can be expressed as the maximum expected value obtained by summing the immediate reward from an action and the anticipated value of the ensuing state \( s' \) resulting from action \( p \)\footnote{For notational simplicity, variables pertaining to the current period are denoted without a prime, while variables representing the subsequent period are indicated with a prime.}. This concept is traditionally encapsulated by Bellman's value function:
\begin{equation}
    V\left(s\right) = \max_{p \in \bm{P}} \left\{ E\left[r_i \mid s, p\right] + \delta E\left[V\left(s'\right) \mid s, p\right] \right\}
\end{equation}

In line with the prevalent methodology for addressing this optimization problem, we adopt the Q-function, which encapsulates the expected reward of taking a particular action in a given state, thereby circumventing the direct solution of Bellman's equations \citep{li2023reinforcement, watkins1989learning}. The Q-function is represented as a matrix with dimensions \( |\bm{S}| \times |\bm{P}| \), and each element is computed according to the following procedure:
\begin{equation}
    Q_j\left(s, p\right) = E\left[r_j \mid s, p\right] + \delta E\left[ \max_{p' \in \bm{P}} Q_j\left(s', p'\right) \mid s, p \right]
\end{equation}

If the true Q-function were known, the agent could readily determine the optimal action to take in the current state. Therefore, the essence of Q-learning lies in approximating the Q-function by exploring various actions while minimizing the loss of rewards due to sub-optimal choices. To estimate the Q matrix \( Q_j \), at each period \( t \), agent \( j \) selects an action \( p_j^t \) based on the current state \( s^t \), observes the resulting state transition \( s^{t+1} \), receives the corresponding reward \( r_j^t \), and updates the Q-value for this state-action pair as follows:
\begin{equation}
    Q_j(s = s^t, p = p_j^t) = (1 - \alpha) Q_j(s, p) + \alpha \left[ r_j^t + \delta \max_{p' \in \bm{P}} Q_j\left(s^{t+1}, p'\right) \right]
\end{equation}

For all other state-action pairs where \( s \neq s^t \) or \( p \neq p_j^t \), the Q-values remain unchanged. Here, we introduce a hyperparameter \( \alpha \), representing the learning rate. This parameter determines the extent to which the updated Q-value depends on newly acquired information during the current period relative to the previous Q-value.

Beginning with minimal information and an arbitrary Q matrix, the agent must learn the value associated with each state-action pair. Given the uncertainty in the environment, particularly the unknown actions of other sellers, each element in the Q table may require numerous iterations to accurately estimate its expected value. However, uniformly exploring all actions can lead to the opportunity cost of selecting sub-optimal actions. Consequently, the Q-learning agent needs to balance exploration and exploitation. To address this, a common strategy is the \( \epsilon \)-greedy approach, where, in each period, the agent selects a random action with probability \( \epsilon \) (exploration) and chooses the optimal action based on the current Q matrix with probability \( 1 - \epsilon \) (exploitation). Moreover, acknowledging that the reliability of the Q matrix improves as the agent gathers more information, it is customary to employ a time-decaying \( \epsilon \) value, thereby increasing the propensity for exploitation over time. In accordance with this methodology, we incorporate an exponential time decay for \( \epsilon \), defined as \( \epsilon = \exp(-\omega t) \), where \( \omega \) denotes the decay rate.

\subsection{Recommender Systems Adopted by the Platform}
\label{subsec: Recsys_main}

In this subsection, we delineate the specific recommender systems employed by the platform. Although our study primarily investigates the economic impact of recommender systems rather than seeking technical enhancements \citep{li2020informative, li2018recommender, fleder2009blockbuster}, we aim to ensure that the algorithm aligns rigorously with the economic setting, as established in prior works on product display algorithms \citep{compiani2024online, derakhshan2022product, ferreira2022learning}. Simultaneously, we also need to maintain the ease of tracking economic properties and conducting counterfactual evaluations.

Achieving these objectives collectively presents inherent complexity, as even minor extensions to the choice model can render the derivation of closed-form expressions for choice probabilities intractable, not to mention the subsequent implications (such as expected revenue or consumer utility) \citep{train2009discrete}. Consequently, even when the platform possesses precise knowledge of the demand model parameters and no heterogeneity exists, determining the optimal assortment and rankings typically necessitates a brute-force approach \citep{compiani2024online}. In our case, the demand model presented in Section \ref{subsec: Economic} incorporates both observed and unobserved heterogeneity, further complicating the derivation of a recommendation strategy. To overcome these challenges, we perform multi-layer numerical integration over the distributions of these heterogeneities and the taste shocks, followed by a brute-force search process to identify the optimal recommendations (including assortment and ranking) based on specific objective functions \citep{compiani2024online}.

As discussed in Section \ref{sec: intro}, previous economic and optimization studies predominantly consider revenue and consumer utility as the primary objective functions for platforms \citep{compiani2024online, derakhshan2022product}. Consequently, we design the recommendation algorithms to optimize these two objectives, aligning with our first research question: \textit{How does the objective of the recommender system influence the algorithmic pricing game?}

Specifically, in each time period, $J$ products are available, each characterized by a vector of static attributes $\bm{X_j}$ and a price $p_j$ set by the seller\footnote{For clarity and simplicity, the time index is omitted in our notation, as previously done in Section \ref{subsec: Economic}.}. Each consumer is represented by observed attributes $\bm{A_i}$, as well as two unobserved attributes, $\beta_i$ and $\tau_i$, which are not observable to the platform. Nevertheless, the platform can infer the distributions of these parameters across the consumer base using available data. In the subsequent subsections, we outline the platform's decision-making process in presenting products to each consumer $i$, focusing on the two aforementioned objectives.

For comparative analysis, we also consider a baseline scenario in which the platform does not utilize any optimization techniques, opting instead to present each consumer with a randomly assorted and ranked selection of products.

\subsubsection{Revenue-Maximization Recommender System}\

The revenue-maximization recommender system aims to maximize the total revenue for all the sellers on the market \citep{compiani2024online}. This approach closely aligns with the objectives of platforms that generate income by collecting a portion of sellers' revenue as commission fees, as observed in prominent platforms such as Amazon and Tmall. 

For each consumer \(i\), the platform is privy to her observable attributes \(\bm{A}_i\) as well as the distributions of unobservable heterogeneities \(\beta_i\) and \(\tau_i\), but not their precise values. Additionally, the exact values of the post-search match values \(\xi_{i,j}\) also remain unknown to the platform; only their distributions are accessible. Consequently, for each potential recommendation set\footnote{For instance, consider three products (indexed as 1, 2, and 3), where the platform must recommend two to the consumer, with the ranking of the displayed products being important. In this case, there are six possible bundles: \{1,2\}, \{2,1\}, \{1,3\}, \{3,1\}, \{2,3\}, \{3,2\}.}, the platform computes the expected revenue and identifies the sets that maximize this expected revenue for exposure through the following process (Algorithm \ref{alg: revenue_max}).

Specifically, the recommendation generation process for consumer \(i\) proceeds as follows. For each potential recommendation set, we first sample the unobserved heterogeneities \(\beta_i\) and \(\tau_i\) (outer loop), followed by sampling the post-search shocks for each product, \(\widehat{\xi_{i,j}}\) (inner loop). In the inner loop, the calibrated structural search model is employed to simulate the consumer's purchase decisions, thereby determining the revenue for each sample. By averaging across the samples in the inner loop, we numerically integrate out the post-search shocks, obtaining the expected revenue conditional on the specific unobservable heterogeneities \((\beta_i\) and \(\tau_i)\). Subsequently, in the outer loop, we average over all sampled unobservable heterogeneities to derive the unconditional expected revenue for the recommendation set. This numerical integration process is widely utilized in optimization and econometrics, such as in simulated maximum likelihood estimation, and will be further detailed in Section \ref{sec: estimation}. Finally, we compare the expected revenues across all recommendation sets, identify those with the highest revenue, and randomly recommend one of these sets to consumer \(i\).

\begin{algorithm}[h!]
\caption{Revenue-Maximization Recommender Systems}
\label{alg: revenue_max}
\begin{algorithmic}[1]
\Require $K, F(\beta_i), F(\tau_i), F(\xi_{i,j}), \bm{A_i}, p_j^t, \text{ and } \bm{X_j} \text{ for } j = 1,2,\ldots,J$
\Require Calibrated Structural Search Model $CM$
\State Derive all the potential $K$-product recommendation sets, each represented by $\bm{B}_b$
\For{each $\bm{B}_b$}
    \For{each sample of unobserved heterogenities (${NS}_1$ samples)}
        \State Sample $\widehat{\beta_i}$ based on $F(\beta_i)$
        \State Sample $\widehat{\tau_i}$ based on $F(\tau_i)$
        \For{each sample of post-search shocks (${NS}_2$ samples)}
            \State Sample $\widehat{\bm{\xi_{i}}}$, which is a vector of $\widehat{\xi_{i,j}}$, based on $F(\xi_{i,j})$
            \State Simulate the purchase decision $\widehat{k_i}$ following $CM$
            \State Derive the platform's award (revenue): $m|(\bm{B}_b, \widehat{\beta_i}, \widehat{\tau_i}, \widehat{\bm{\xi_{i}}}) = p_{\widehat{k_i}}^t$
        \EndFor
        \State Approximate the conditional expected revenue: 
        $$
        m|(\bm{B}_b, \widehat{\beta_i}, \widehat{\tau_i}) = \frac{1}{{NS}_2} \sum m|(\bm{B}_b, \widehat{\beta_i}, \widehat{\tau_i}, \widehat{\bm{\xi_{i}}})
        $$
    \EndFor
    \State Approximate the conditional expected revenue: 
    $$
    m|\bm{B}_b = \frac{1}{{NS}_1} \sum m|(\bm{B}_b, \widehat{\beta_i}, \widehat{\tau_i})
    $$
\EndFor
\State Find the $\bm{B}_b$ with the largest expected revenue $m|\bm{B}_b$
\State Output: Equally assign the exposure to these sets with the largest expected revenue
\end{algorithmic}
\end{algorithm}

\subsubsection{Utility-Maximization recommender system}\

In contrast, utility-maximization recommender systems are designed to enhance the utility experienced by consumers on the platform \citep{compiani2024online}. Utility serves as an economic measure of consumer satisfaction, which benefits the platform by improving consumer retention rates, thereby making it a common objective for many platforms \citep{zhang2019deep, ursu2018power}. Similar to revenue-maximization recommender systems, utility-maximization systems also employ numerical integration techniques to estimate the expected consumer utility for each potential recommendation set. The primary distinction lies in the post-simulation analysis: for each sample, the calibrated structural model is used to simulate consumer behavior, after which the search history and purchase decisions are utilized to calculate utility (purchase utility minus total search costs) instead of revenue. Once the expected utility for each recommendation set is derived, the platform identifies those sets that maximize utility and randomly selects one of these optimal sets to recommend to consumer \(i\). This process is further detailed in Algorithm \ref{alg: utility_max}.

\begin{algorithm}[h!]
\caption{Utility-Maximization Recommender Systems}
\label{alg: utility_max}
\begin{algorithmic}[1]
\Require $K, F(\beta_i), F(\tau_i), F(\xi_{i,j}), \bm{A_i}, p_j^t, \text{ and } \bm{X_j} \text{ for } j = 1,2,\ldots,J$
\Require Calibrated Structural Search Model $CM$
\State Derive all the potential $K$-product recommendation sets, each represented by $\bm{B}_b$
\For{each $\bm{B}_b$}
    \For{each sample of unobserved heterogenities (${NS}_1$ samples)}
        \State Sample $\widehat{\beta_i}$ based on $F(\beta_i)$
        \State Sample $\widehat{\tau_i}$ based on $F(\tau_i)$
        \For{each sample of post-search shocks (${NS}_2$ samples)}
            \State Sample $\widehat{\bm{\xi_{i}}}$, which is a vector of $\widehat{\xi_{i,j}}$, based on $F(\xi_{i,j})$
            \State Simulate the search sequence $\widehat{\bm{H_i}}$ and purchase decision $\widehat{k_i}$ following $CM$
            \State Derive the utility: 
            $$
            m|(\bm{B}_b, \widehat{\beta_i}, \widehat{\tau_i}, \widehat{\bm{\xi_{i}}}) = u_{i,\widehat{k_i}} - \sum_{j \in \widehat{\bm{H_i}}} c_{i,j}
            $$
        \EndFor
        \State Approximate the conditional expected utility: 
        $$
        m|(\bm{B}_b, \widehat{\beta_i}, \widehat{\tau_i}) = \frac{1}{{NS}_2} \sum m|(\bm{B}_b, \widehat{\beta_i}, \widehat{\tau_i}, \widehat{\bm{\xi_{i}}})
        $$
    \EndFor
    \State Approximate the conditional expected utility: 
    $$
    m|\bm{B}_b = \frac{1}{{NS}_1} \sum m|(\bm{B}_b, \widehat{\beta_i}, \widehat{\tau_i})
    $$
\EndFor
\State Find the $\bm{B}_b$ with the largest expected utility $m|\bm{B}_b$
\State Output: Equally assign the exposure to these sets with the largest expected utility
\end{algorithmic}
\end{algorithm}

\section{Data, Structural Estimation, and Parameterization}
\label{sec: estimation}

Although we aim to design our framework to be robust against parameter variations and minimize reliance on assumptions, certain parameters still need to be specified prior to conducting numerical experiments. In this section, we describe the calibration of the structural search model using real-world data, the setup of the pricing algorithms, and the integration of these components for step-by-step simulations.

\subsection{Parameterization of the Consumer Demand Model}

In modeling consumer demand, we aim to derive consumers' price sensitivity, attribute preferences, search costs, and positional biases within a \textit{real-world} context. To achieve this, we draw upon established literature and estimate our model using customer search and choice data made publicly available by Expedia \citep{compiani2024online, ursu2018power}.

\subsubsection{Data} \

The Expedia dataset is provided at the level of \emph{search impressions}, defined as ordered lists of hotels and their characteristics presented to consumers in response to their trip queries. Notably, this dataset originates from a field experiment wherein consumers searching for hotels during the observational window were randomized into two groups: (i) those viewing Expedia's standard ranking, where hotels are ordered by relevance, and (ii) those viewing a random ranking, where hotels are displayed in a random order. Consequently, for the randomized group, the hotel display is exogenous to their attributes, eliminating the influence of the platform's existing algorithms on the customer modeling process. We utilize this random-ranking sample for our model estimation \citep{chung2024simulated, ursu2018power}.

Following prior literature, we also partition the dataset by travel destination, recognizing that each destination may differ significantly in terms of hotels and travelers, and that hotels in different destinations do not directly compete with each other \citep{chung2024simulated, ursu2018power}. We thus focus on the destination with the \emph{largest} number of observations (destination ID 8192). After applying additional filtering procedures to exclude null values and outliers, we obtain 35,769 observations, where each observation represents one available alternative. Further details on data cleaning, variable descriptions, and summary statistics are provided in Appendix \ref{appendix:data}.

\subsubsection{Estimation} \

As researchers, we observe some covariates ($\bm{A_i}$, $\bm{X_j}$), consumers' search sequences $\bm{H_i} = \{\bm{H_i}(0), \ldots, \bm{H_i}(h)\}$, and their final purchase decisions $k$. However, the utility parameters ($\beta_i$, $\bm{\Gamma}$, $\bm{\Theta}$) and search cost parameters ($\tau_i$, $\varphi$) are unobserved and need to be estimated from the data. Following the sequential search literature, we also do not observe the exact values of the post-search shocks $\xi_{i,j}$ but assume they are drawn from the standard normal distribution $N(0,1)$ \citep{ursu2018power, weitzman1979optimal}\footnote{This is a standard assumption in the sequential search literature because, in any utility maximization framework, only differences among choices matter \citep{ursu2023sequential}. Consequently, only the \textit{relative} magnitudes of the search cost and the post-search variance are crucial in balancing the costs and benefits of conducting searches. This normalization allows us to fix one part while estimating the other.}. From the consumers' decision rules, we can infer the following mathematical relationships:
\begin{enumerate}
    \item A product searched later has a reservation utility higher than the utilities of products searched earlier:
    \begin{equation}
    z_{i,\bm{H_i}(q)} \geq \max_{j=0}^{q-1} u_{i,\bm{H_i}(j)}, \quad \forall q \in \{1, \dots, h\}.
    \end{equation}
    \item The reservation utilities of all unsearched products are lower than the maximum utility among the searched ones:
    \begin{equation}
    \max_{q \in \bm{B_i} \setminus \bm{H_i}} z_{i,\bm{H_i}(q)} \leq \max_{j \in \bm{H_i}} u_{i,j}.
    \end{equation}
    \item The purchased product has the highest utility among all searched products:
    \begin{equation}
    u_{i,k} \geq \max_{j \in \bm{H_i}} u_{i,j}.
    \end{equation}
\end{enumerate}

By integrating these relationships, the probability of consumers' search sequences and purchase decisions that align with the observed data can be expressed as:
\begin{equation}
\begin{aligned}
P_{i,k,\bm{H_i}} = \Pr &\left( z_{i,\bm{H_i}(q)} \geq \max_{j=0}^{q-1} u_{i,\bm{H_i}(j)}, \quad \forall q \in \{1, \dots, h\} \right) \cap \\
&\left( \max_{q \in \bm{B_i} \setminus \bm{H_i}} z_{i,\bm{H_i}(q)} \leq \max_{j \in \bm{H_i}} u_{i,j} \right) \cap \\
&\left( u_{i,k} \geq \max_{j \in \bm{H_i}} u_{i,j} \right)
\end{aligned}
\end{equation}

To estimate the model, we address two primary challenges. First, structural search models typically lack closed-form expressions that directly relate parameters to choice probabilities. Consequently, consistent with prior literature, we employ simulated maximum likelihood (SML) estimation using a logit-smoothed accept-reject (AR) simulator \citep{ursu2023sequential}. Second, to enhance the model's flexibility in analyzing the pricing game under the influence of recommendations, it is essential to account for unobservable heterogeneities in price coefficients and search costs. Specifically, we assume that the price coefficient and search cost are drawn from log-normal distributions (see Equations \ref{equ:beta} and \ref{equ:c}): 
\begin{align}
    & \beta_i \sim \exp(m_{\beta} + \sigma_{\beta} \cdot y_{\beta,i}) \label{equ:beta} \\
    & c_{i,j} \sim \exp(m_{\tau} + \sigma_{\tau} \cdot y_{\tau,i} + \varphi \cdot pos_{i,j}) \label{equ:c}
\end{align}
where $y_{\beta,i} \sim N(0,1)$ and $y_{\tau,i} \sim N(0,1)$.

Although SML is commonly utilized for econometric models with random coefficients \citep{train2009discrete}, incorporating an additional layer of simulation to capture unobservable heterogeneities substantially increases the computational burden. To mitigate this, we innovatively integrate Gaussian Hermite quadrature into the estimation procedure \citep{akcsin2013structural}. To the best of our knowledge, this represents the first application of the Hermite approach to estimate a structural search model incorporating unobservable heterogeneities. After addressing these challenges, we obtain the model estimates presented in Table~\ref{tab:table_estimation}. 

\begin{table}[ht]
\centering
\caption{Structural Estimation Results}
\label{tab:table_estimation}
\newcolumntype{L}[1]{>{\raggedright\arraybackslash}p{#1}}
\newcolumntype{C}[1]{>{\centering\arraybackslash}p{#1}}
\newcolumntype{R}[1]{>{\raggedleft\arraybackslash}p{#1}}
\renewcommand\arraystretch{1.5} 
\begin{tabular}{L{3cm}L{3cm}L{2cm}}
\hline 
\hline
 \textbf{Variable} & \textbf{Coefficients} & \textbf{Std}
\\
 $m_{\beta}$ & -1.0384$^{***}$ & (0.0047)\\
 $\sigma_{\beta}$ & 0.0067$^{***}$ & (0.0025)\\
 $\Gamma_{HotelStars}$ & 0.4955$^{***}$ & (0.0197)\\
 $\Gamma_{HotelReview}$ & -0.1897$^{***}$ & (0.0387)\\
 $\Gamma_{Brand}$ & 0.0030 & (0.0235)\\
 $\Gamma_{HotelLocationScore}$ & -0.1851$^{***}$ & (0.0469)\\
 $\Gamma_{Promo}$ & 0.0408$^{*}$ & (0.0247)\\
 $\Theta_{ReqRoomNum}$ & -0.1315$^{***}$ & (0.0197)\\
 $\Theta_{SatNight}$ & 0.1045$^{***}$ & (0.0168)\\
 $v_{i,0}$ & 0.1653 & (0.3173)\\
 $m_{\tau}$ & -1.1490$^{***}$ & (0.0172) \\
 $\sigma_{\tau}$ & 0.0050$^{*}$ & (0.0029) \\
 $\varphi$ & 0.0050$^{***}$ & (0.0009) \\ 
\hline \hline
\end{tabular}
\par
\vspace{1em}
\begin{minipage}{0.8\textwidth}
\centering
{\footnotesize  Note: Bootstrap standard errors in parentheses; $^{***}$ p$<$0.01, $^{**}$ p$<$0.05, $^{*}$ p$<$0.1.\par}
\end{minipage}
\end{table}

\subsection{Parameterization of the Pricing Algorithms}

In our repeated game framework, the pricing algorithms necessitate the specification of several hyperparameters prior to simulation. Since these parameters do not influence the qualitative outcomes, we adopt the hyperparameter settings from \cite{calvano2020artificial}, specifically assigning a learning rate of \(\alpha = 0.15\) and a discount factor of \(\delta = 0.95\). To accommodate the presence of three agents, we enhance exploration by utilizing a smaller exploration rate, \(\omega = 1.5 \times 10^{-6}\). We define the minimum and maximum permissible prices for sellers as the 10th and 90th percentiles of the dataset's prices (i.e., 0.5 and 2.59), respectively. This price range is discretized into nine equal intervals, resulting in ten available pricing actions. Additionally, we relax these parameter assumptions in Section \ref{sec: Extended}.

\subsection{Integration into the Simulation}

To evaluate the outcomes and address our research questions within the repeated game framework, we implement several additional steps. First, for our primary simulation, we consider a scenario involving three sellers, where the platform recommends two out of the three available products to each consumer in each period. This configuration represents the most fundamental case in which assortment and position biases are relevant, while avoiding excessive computational complexity following prior studies \citep{dou2024ai, calvano2020artificial}. 

Second, to establish the static characteristics of the sellers, we utilize the median values from the dataset, thereby creating a balanced market that facilitates easy interpretations of effect sizes. However, this approach results in homogeneity among alternatives aside from price. To introduce horizontal differentiation, we add a zero-mean term of the form $(\mu \cdot N(0,1))$, where $\mu$ can be adjusted to simulate various economic environments and address our third research question\footnote{Since our utility magnitudes are comparable to those in \cite{calvano2020artificial}, we also vary $\mu$ from 0 to 0.5 to sufficiently explore its impact and employ $\mu = 0.25$ to illustrate the main results.}.

Third, we employ the empirical distribution of all consumers from the dataset. This distribution, combined with the calibrated structural search model, enables us to generate demand and thereby determine the rewards for both sellers and the platform. To compute the expected value of these rewards, we follow the same numerical integration procedure used for the recommender system, accounting for both consumer heterogeneity and taste shocks. Having established these elements, we integrate the calibrated structural search model and the pricing algorithms into the repeated game framework to carry out the simulations.

\section{Results}
\label{sec: results}

After completing these aforementioned steps, we proceed to conduct numerical experiments to address our research questions. Specifically, as outlined in Algorithm \ref{alg1}, each simulation run extends over multiple periods. At the commencement of each period, each seller utilizes its respective pricing algorithm, as detailed in Section \ref{subsec: Q-learning}, to determine pricing decisions. Subsequently, the platform generates recommendations for each consumer following the strategy described in Section \ref{subsec: Recsys_main}. Consumers then observe the recommended set and make purchase decisions based on the calibrated structural search model presented in Section \ref{subsec: Economic}. Each seller realizes revenue and updates its algorithm accordingly. With parameters fixed as described in Section \ref{sec: estimation}, the simulation framework is fully specified.

In the subsequent three subsections, we vary different elements of the framework to evaluate how these variations produce different counterfactual outcomes, thereby systematically addressing each of our research questions.

\subsection{Effect of the Recommender Systems' Objectives}
\label{sec:base_res}

As motivated in our \textit{RQ1}, one of the fundamental economic aspects of recommender systems is their underlying objective, as it directly shapes which products are highlighted for consumers and influences sellers’ rewards and game directions. Among the commonly considered objectives are \textit{Revenue-Maximization}, which seeks to increase total platform-wide revenue and thereby aligns with the platform’s economic interests through commission fees, and \textit{Utility-Maximization}, which aims to enhance consumer utility, potentially benefiting the platform by fostering consumer retention \citep{compiani2024online, feldman2022customer, derakhshan2022product, ursu2018power}.

To evaluate how different objectives influence the dynamics of the algorithmic game, we conduct a series of numerical experiments. Specifically, we compare the equilibrium outcomes of the game (prices, overall revenue, and consumer utility) under three distinct scenarios as specified in Section \ref{subsec: Recsys_main}: (i) a revenue-maximization recommender system, (ii) a utility-maximization recommender system, and (iii) a baseline scenario that does not incorporate recommendation optimization as a point of reference.

\begin{figure}[h]
\centering
\subfloat[Revenue-Maximization RS]{\includegraphics[width=.5\linewidth]{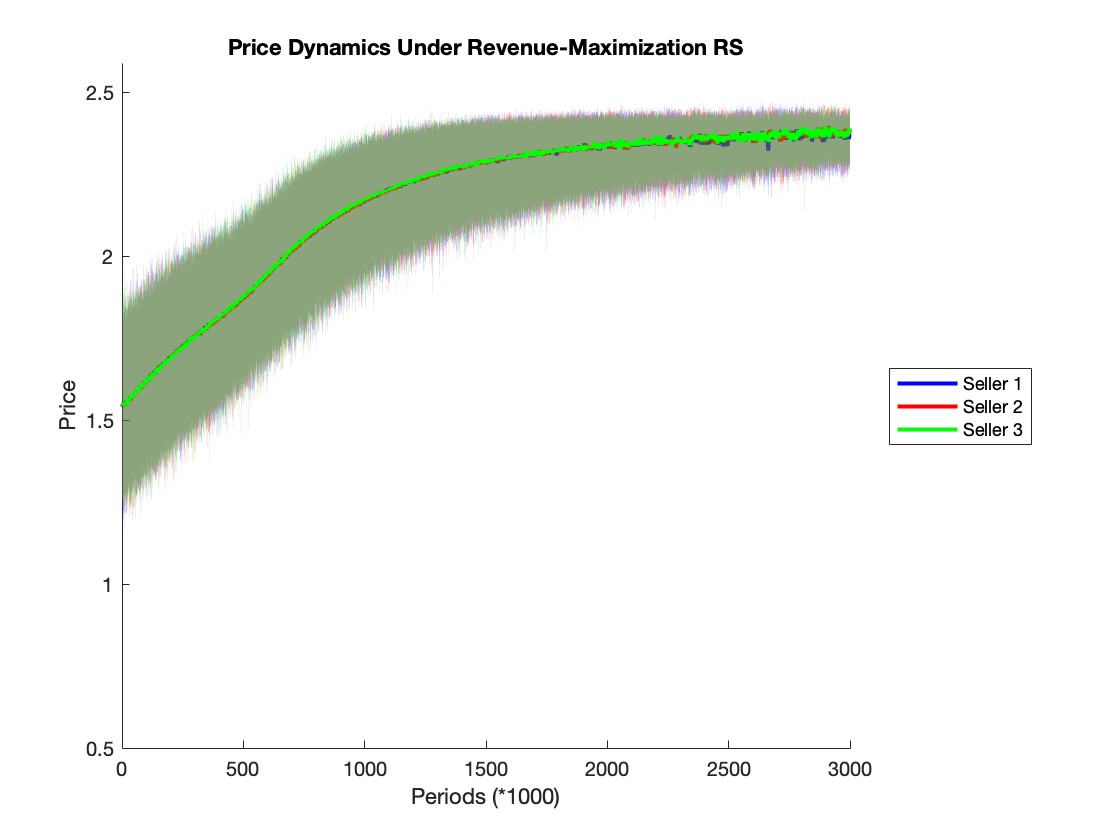}}\hfill
\subfloat[Utility-Maximization RS]{\includegraphics[width=.5\linewidth]{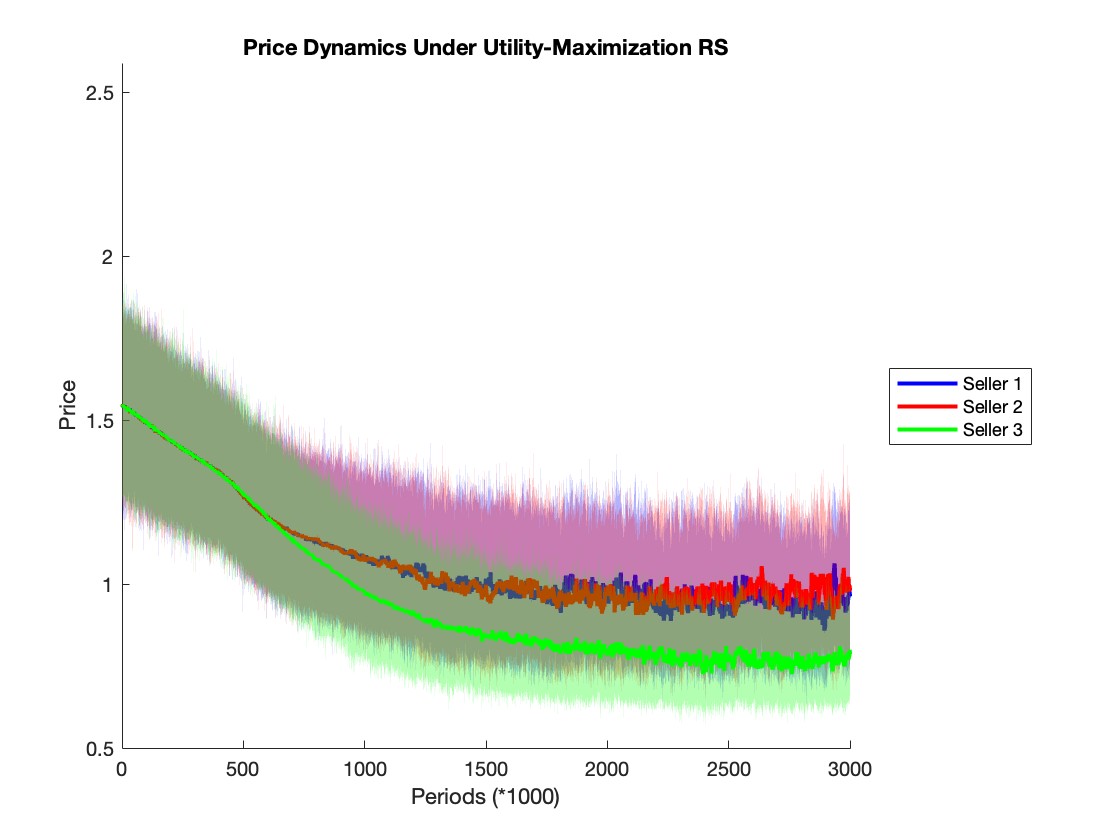}}\par 
\subfloat[Baseline]{\includegraphics[width=.5\linewidth]{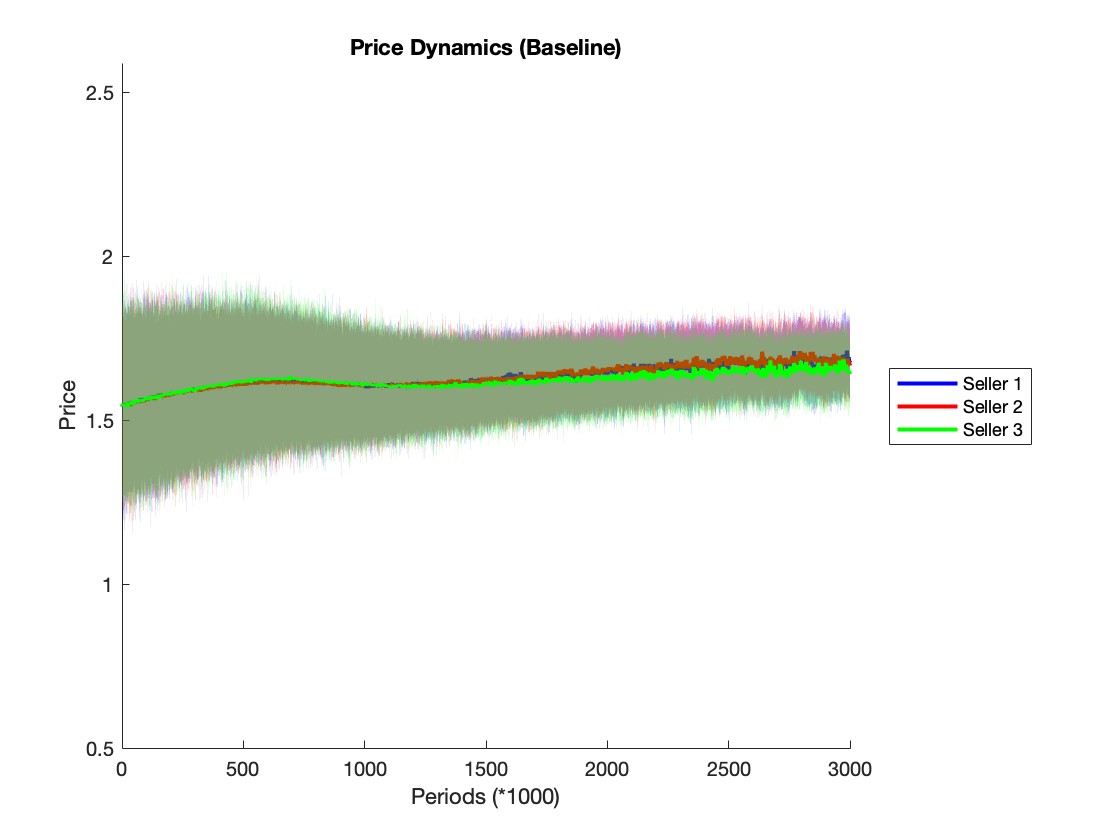}}\\
\caption{Learning Dynamics}
\medskip
\begin{minipage}{0.8\textwidth}
{\footnotesize  Note: The points on the continuous lines depict the moving average, with a window size of 1000, of the average price observed across 50 experiments at each $t$. The shaded region's upper and lower limits indicate the highest and lowest values occurring within the window.\par}
\end{minipage}
\label{fig:learn_base}
\end{figure}

For each scenario, we conduct 50 simulation runs, each spanning 3,000,000 periods\footnote{Q-learning agents often converge slowly because only one element is updated in each period \citep{calvano2020artificial}. However, as shown in Figure \ref{fig:learn_base}, the game's duration primarily affects the magnitude of the effects rather than their qualitative direction.}. At each period, we record the average price and revenue for the three sellers, as well as the average consumer utility across all consumers. These values are then averaged over the 50 runs, and their means from the final 1,000 periods are reported in Table \ref{tab:table_base}. Standard deviations, presented in parentheses, provide insight into the degree of end-game fluctuation.

\begin{table}[ht]
\centering
\caption{Price, Revenue, and Utility Comparison Across Objectives}
\label{tab:table_base}
\newcolumntype{L}[1]{>{\raggedright\arraybackslash}p{#1}}
\newcolumntype{C}[1]{>{\centering\arraybackslash}p{#1}}
\newcolumntype{R}[1]{>{\raggedleft\arraybackslash}p{#1}}
\renewcommand\arraystretch{1.5} 
\begin{tabular}{L{1.5cm}C{4cm}C{4cm}C{4cm}}
\hline 
\hline
 \textbf{ } & \textbf{Revenue-Maximization} & \textbf{Utility-Maximization} & \textbf{Baseline} 
\\
 \hline
 Prices & 2.3844 (0.0137) & 0.9095 (0.0478) & 1.6617 (0.0236) \\
 Revenue & 0.2583 (0.0003) & 0.1174 (0.0041) & 0.2163 (0.0015) \\
 Utility & 0.3749 (0.0012) & 0.6506 (0.0063) & 0.4765 (0.0032) \\
\hline \hline
\end{tabular}
\end{table}

Our findings indicate that the objectives of recommender systems substantially shape the trajectory of pricing strategies and their eventual equilibrium outcomes. Under revenue-maximization, sellers gravitate toward more collusive behavior, resulting in higher equilibrium prices (2.3844) and revenue (0.2583) at the expense of consumer utility (0.3749). Conversely, when the system aims to maximize consumer utility, the equilibrium is less collusive, featuring lower prices (0.9095) and higher consumer utility (0.6506), but reduced seller revenue (0.1174).

Figure \ref{fig:learn_base} illustrates the temporal evolution of prices under the three scenarios, further corroborating our findings. In the revenue-maximization setting, sellers gradually and stably converge to higher prices. Under utility-maximization, they tend to settle at lower prices but exhibit greater volatility in their pricing behavior.

An additional noteworthy pattern emerges under utility-maximization recommender systems: two sellers maintain slightly higher prices, while one sets a lower price. In this asymmetric equilibrium, each seller seeks to optimize its own revenue. The two high-price sellers face lower exposure and sales, yet each sale generates relatively higher revenue. The low-price seller benefits from increased exposure and sales volume, but at a lower per-sale revenue. This interplay ensures that all three sellers achieve similar revenue levels, while the platform also meets its utility-maximization objective despite the asymmetric pricing configuration.

To further elucidate this asymmetry, we can examine the underlying incentives of the sellers and the platform. On one hand, a high-priced seller might find it beneficial to deviate by lowering its price. However, after that, in the presence of two low-priced competitors, the remaining high-priced seller receives minimal exposure and revenue due to the platform’s utility-maximizing recommender system. This scenario incentivizes the high-priced seller to also reduce its price. The resulting intensified competition diminishes the revenues of all three sellers, rendering such deviations unstable. Conversely, sellers have limited motivation to increase prices, as doing so would lead to decreased exposure and revenue. These opposing dynamics converge to an asymmetric equilibrium that exhibits significant fluctuations, thereby accounting for the relatively high standard deviation observed under the utility-maximization framework.

Conversely, the revenue-maximization scenario converges rapidly and maintains relative stability. In this context, both the sellers and the platform are incentivized to sustain higher prices. Starting from a lower price, if a seller raises its price, the likelihood of purchase will decrease, assuming all other factors remain constant. However, in the presence of revenue-maximizing recommender systems, this reduced likelihood of purchase may be offset by increased product exposure, thereby compensating for the higher price. As a result, prices within the revenue-maximization scenario can escalate to collusive levels and remain at these levels over time. At these collusive price points, all products within the recommendation set receive equal exposure. If a seller deviates by setting a different price, this deviation leads to a decrease in overall revenue for displaying the deviating seller. Consequently, the platform responds by prioritizing the other sellers: the deviating seller experiences significantly lower exposure and rewards, while the non-deviating sellers receive even higher rewards. This mechanism incentivizes the deviating seller to revert to the collusive price, and the non-deviating sellers have no incentive to follow the deviation. Consequently, driven by the revenue-driven incentives of the three sellers and the platform's revenue-maximizing recommendation system, a collusive price equilibrium is both achievable and stable.

Furthermore, certain recommender systems in two-sided markets balance the interests of various stakeholders by adopting a composite objective that integrates both revenue and consumer utility. Consequently, beyond the aforementioned analysis, we examine the impact of varying the relative weights assigned to revenue and utility objectives on market outcomes, as detailed in Appendix \ref{appendix:optweight}. The findings indicate a monotonic relationship between these weights and the resulting equilibrium, thereby highlighting the critical role of recommender system objectives in shaping algorithmic pricing dynamics. Therefore, when antitrust regulators aim to control supracompetitive prices, they should not solely attribute such phenomena to sellers and pricing algorithm providers. Instead, platforms that implement recommender systems should also be considered as part of the responsible entities.

\subsection{Effect of the Recommendation Size}

Apart from the objectives of recommender systems, another aspect of particular interest regarding their economic impact is the size of recommendations. As motivated in our second research question (\textit{RQ2}), this dimension holds significant economic importance, as it directly influences search friction and competition among sellers \citep{moraga2023consumer, hong2006using}. Guided by the antitrust and algorithmic collusion literature’s emphasis on enhancing consumer welfare \citep{baker2019antitrust}, it is natural to question whether displaying more products can effectively achieve this goal. The rationale is as follows: with a greater number of options, consumers can disregard products they do not prefer while selecting those that enhance their utility. This intuition is supported by choice modeling literature, where the welfare function of a product set increases as more products are included \citep{train2009discrete}.

To assess the impact of recommendation size when using the utility-maximization recommender system and address our \textit{RQ2}, we conduct numerical experiments with $K = 1, 2, 3$ under Top-$K$ utility maximization recommender systems. For each scenario, we perform 50 simulation runs (each consisting of 3,000,000 periods) and record the average equilibrium price, revenue, and utility in Table \ref{tab:table_size}.

Contrary to conventional wisdom and the choice modeling literature, as illustrated in the third row of Table \ref{tab:table_size}, increasing the number of displayed products does not consistently enhance consumer utility ($0.6506 > 0.6128 > 0.5522$). This seemingly counterintuitive result arises from two competing mechanisms. On the one hand, displaying more products provides consumers with greater opportunities to identify products that offer higher utility, potentially increasing their overall utility. On the other hand, a larger number of displayed products diminishes the platform’s ability to effectively influence prices through recommendations. High-priced products continue to receive exposure, and some consumers purchase them based on their horizontal preferences. Consequently, higher-priced products experience relatively smaller losses in their rewards, leading to relatively higher equilibrium prices after repeated interactions. This effect is reflected in the first row of Table \ref{tab:table_size} ($1.3115 > 0.9095 > 0.6857$). When the latter mechanism predominates, recommending more products can be detrimental from the consumer's perspective.

This result also suggests that future research on the economics of recommender systems should also consider sellers' price responses as a crucial outcome, particularly in the presence of pricing algorithms in the market.

\begin{table}[ht]
\centering
\caption{Price, Revenue, and Utility Comparison Across Recommendation Sizes}
\label{tab:table_size}
\newcolumntype{L}[1]{>{\raggedright\arraybackslash}p{#1}}
\newcolumntype{C}[1]{>{\centering\arraybackslash}p{#1}}
\newcolumntype{R}[1]{>{\raggedleft\arraybackslash}p{#1}}
\renewcommand\arraystretch{1.5} 
\begin{tabular}{L{1.5cm}C{4cm}C{4cm}C{4cm}}
\hline 
\hline
 \textbf{ } & \textbf{K = 1} & \textbf{K = 2} & \textbf{K = 3} 
\\
 \hline
 Prices & 0.6857 (0.0295) & 0.9095 (0.0478) & 1.3115 (0.0261) \\
 Revenue & 0.0788 (0.0014) & 0.1174 (0.0041) & 0.2040 (0.0027) \\
 Utility & 0.5522 (0.0022) & 0.6506 (0.0063) & 0.6128 (0.0045) \\
\hline \hline
\end{tabular}
\end{table}

\subsection{Horizontal Differentiation Levels as the Moderator}

Building on the mechanisms discussed in \textit{RQ2}, the extent of horizontal differentiation appears to be a critical moderator influencing the benefits of displaying additional products. As highlighted in our motivation process of \textit{RQ3}, when products are highly homogeneous aside from their prices, most consumers simply choose the least expensive option and ignore others. Consequently, increasing the number of displayed products provides minimal benefit under conditions of low horizontal differentiation. Given the pivotal role that horizontal differentiation plays in consumer decision-making, prior literature in price competition and industrial organization consistently identifies it as an important factor \citep{colombo2013product, thomadsen2007costly, tyagi1999relationship}. Within our pricing game framework, similar logic suggests that greater horizontal differentiation could enhance consumer utility by offering a wider range of product choices.

To investigate the moderating role of horizontal differentiation and thereby address \textit{RQ3}, we conduct a series of simulations by varying $\mu$ and examining scenarios where $K = 1, 2, 3$. We replicate the same analyses as in the previous subsection, demonstrating two sets of results below: one with $\mu = 0$ in Table \ref{tab:table_size_mu0} and another with $\mu = 0.5$ in Table \ref{tab:table_size_mu05}.

Surprisingly, our findings uncover a counterintuitive pattern. As horizontal differentiation increases, the previously observed ``more is less'' phenomenon emerges (see row 3 of Table \ref{tab:table_size_mu05}). In contrast, when horizontal differentiation is relatively low, increasing the number of recommended products actually enhances consumer utility (see row 3 of Table \ref{tab:table_size_mu0}).

This outcome can be explained as follows. On the one hand, greater horizontal differentiation strengthens the mechanism by which consumers more effectively identify products that align with their preferences, thus gaining additional utility from a broader range of options. On the other hand, it reinforces a secondary mechanism: as horizontal differentiation increases, consumers' choices become more preference-driven rather than price-driven. Consequently, when the recommendation set is large, higher-priced products experience relatively smaller declines in rewards and can sustain elevated prices due to persistent demand fueled by horizontal preferences. 

When horizontal differentiation is high, both mechanisms intensify, but the latter may dominate. As a result, contrary to conventional expectations, displaying more products can sometimes reduce consumer utility when horizontal differentiation is high, while it increases consumer utility when horizontal differentiation is low.

\begin{table}[ht]
\centering
\caption{Price, Revenue, and Utility Comparison Across Recommendation Sizes ($\mu = 0$)}
\label{tab:table_size_mu0}
\newcolumntype{L}[1]{>{\raggedright\arraybackslash}p{#1}}
\newcolumntype{C}[1]{>{\centering\arraybackslash}p{#1}}
\newcolumntype{R}[1]{>{\raggedleft\arraybackslash}p{#1}}
\renewcommand\arraystretch{1.5} 
\begin{tabular}{L{1.5cm}C{4cm}C{4cm}C{4cm}}
\hline 
\hline
 \textbf{ } & \textbf{K = 1} & \textbf{K = 2} & \textbf{K = 3} 
\\
 \hline
 Prices & 0.6368 (0.0241) & 0.8837 (0.0498) & 0.9386 (0.0459) \\
 Revenue & 0.0765 (0.0013) & 0.1036 (0.0035) & 0.1257 (0.0041) \\
 Utility & 0.5356 (0.0019) & 0.6183 (0.0056) & 0.6356 (0.0067) \\
\hline \hline
\end{tabular}
\end{table}

\begin{table}[ht]
\centering
\caption{Price, Revenue, and Utility Comparison Across Recommendation Sizes ($\mu = 0.5$)}
\label{tab:table_size_mu05}
\newcolumntype{L}[1]{>{\raggedright\arraybackslash}p{#1}}
\newcolumntype{C}[1]{>{\centering\arraybackslash}p{#1}}
\newcolumntype{R}[1]{>{\raggedleft\arraybackslash}p{#1}}
\renewcommand\arraystretch{1.5} 
\begin{tabular}{L{1.5cm}C{4cm}C{4cm}C{4cm}}
\hline 
\hline
 \textbf{ } & \textbf{K = 1} & \textbf{K = 2} & \textbf{K = 3} 
\\
 \hline
 Prices & 0.6887 (0.0261) & 0.8781 (0.0410) & 1.7614 (0.0283) \\
 Revenue & 0.0815 (0.0016) & 0.1292 (0.0047) & 0.2740 (0.0028) \\
 Utility & 0.6102 (0.0026) & 0.7471 (0.0066) & 0.6569 (0.0050) \\
\hline \hline
\end{tabular}
\end{table}

\section{Framework Comparison and Extended Analyses}
\label{sec: Extended}

To incorporate recommender systems and address our research questions, we develop a repeated game framework that builds upon, yet significantly differs from, the framework proposed by \cite{calvano2020artificial}. We nonetheless adopt several assumptions from \cite{calvano2020artificial}, and thus conduct a series of extended analyses to examine these assumptions individually, thereby providing a more holistic perspective. All comparisons and extended analyses are summarized in Table \ref{tab:compare}.

\begin{table}[ht]
\centering 
\caption{Framework Comparison and Summary of Extensions}
\label{tab:compare}
\newcolumntype{L}[1]{>{\raggedright\arraybackslash}p{#1}}
\newcolumntype{C}[1]{>{\centering\arraybackslash}p{#1}}
\newcolumntype{R}[1]{>{\raggedleft\arraybackslash}p{#1}}
\renewcommand\arraystretch{1} 
\scalebox{0.9}{
\begin{tabular}{L{5.5cm} L{5cm} L{5.7cm}}
\\[-1.8ex]\hline  
\hline \\[-1.8ex]  
   & \cite{calvano2020artificial} & This Work \\
\hline \\[-1.8ex] 
\textit{Demand Model: Model Form} & Reduced-form Models & Structural Models (more in Appendix \ref{appendix:visualization}, \ref{appendix:logit}, \ref{appendix:heterologit})  \\ 
\textit{Demand Model: Data Usage} & No data & Real-world dataset (more in Appendix \ref{appendix:data})\\ 
\textit{Demand Model: Choice Set} & All the products & Recommended Products \\ 
\textit{Demand Model: Position Bias} & Not Applicable & Yes \\ 
\textit{Pricing Algorithm: Sufficient Exploration} & Yes & Relaxed (more in Appendix \ref{appendix:explore_more})\\ 
\textit{Pricing Algorithm: Forward-Looking} & Yes & Relaxed (more in Appendix \ref{appendix:mab}) \\ 
\textit{Pricing Algorithm: Objective} & Revenue & Relaxed (more in Appendix \ref{appendix:pricingobj}) \\ 
\textit{Game Sequence} & Simultaneous & Relaxed (more in Appendix \ref{appendix:asyn}) \\ 
\textit{Recommender Systems} & No & Yes (More in Appendix \ref{appendix:optdemand} and \ref{appendix:optweight}) \\ 
\textit{Objective Misalignment} & Not Applicable & Yes \\ 
\hline \\[-1.8ex] 
\end{tabular}
}
\end{table}

With respect to the demand model, most studies on algorithmic collusion follow \cite{calvano2020artificial} by employing a reduced-form model in which parameters are directly specified to generate demand responses. In contrast, our main analyses seek to more accurately capture consumer behavior in the presence of recommended products and to systematically characterize the underlying utility functions. To this end, we develop a sequential search model that incorporates heterogeneous consumer preferences and search costs, as detailed in Section \ref{subsec: Economic} and Appendix \ref{appendix:visualization}. We estimate the model parameters using the procedure outlined in Section \ref{sec: estimation} and a real-world dataset described in Appendix \ref{appendix:data}, enabling us to conduct counterfactual simulations and derive empirically grounded insights. Further distinguishing our approach, \cite{calvano2020artificial} assume that all products are fully visible to consumers, providing them with complete information. In contrast, and to more closely reflect the information overload often encountered in online markets, our model restricts the consumer choice set primarily to recommended products and introduces search costs for additional information. We also incorporate position bias into the sequential search model, allowing for differential consideration of products based on their placement within the recommendation set. Although this sequential search model offers several advantages, we additionally employ a canonical logit demand model for robustness checks. As shown in Appendices \ref{appendix:logit} and \ref{appendix:heterologit}, the core insights remain robust even when using a logit demand model with or without heterogeneity in the outside option.

For the pricing algorithm, we follow \cite{calvano2020artificial} and implement a basic Q-learning approach. Q-learning is chosen due to its prominence in reinforcement learning, its minimal input requirements, and its nonparametric simplicity. However, Q-learning inherently embodies a forward-looking component through its state transitions. Therefore, we conduct a robustness check by relaxing the forward-looking assumption and modeling the algorithms as basic multi-armed bandits. As demonstrated in Appendix \ref{appendix:mab}, this modification does not fundamentally alter our results, indicating that forward-looking behavior is not essential for the core mechanism underlying our findings. Moreover, we also examine alternative seller objectives in Appendix \ref{appendix:pricingobj} to explore the broader applicability of our findings.

Regarding the sequence of the game, \cite{calvano2020artificial} assume that all sellers make their decisions simultaneously. While we retain this assumption in our primary analyses, we also relax it by allowing sellers to update their prices asynchronously. As shown in Appendix \ref{appendix:asyn}, the core results remain stable under both simultaneous and asynchronous updating protocols, reinforcing the generality of our conclusions.

Finally, our framework introduces a key innovation by explicitly incorporating recommender systems. Beyond addressing the technical challenges of developing an algorithm that is both empirically relevant and theoretically tractable, this addition enables us to examine objective misalignments among various stakeholders. In the main analyses, we illustrate the impact of differing objectives by considering two polar cases. To further enhance realism, we also examine scenarios in which the platform optimizes a weighted combination of revenue and utility, with results reported in Appendix \ref{appendix:optweight}. Additionally, although consumer utility is closely related to purchase probability and demand, making demand-maximization theoretically similar to utility-maximization, we nevertheless explicitly test this scenario in Appendix \ref{appendix:optdemand}. In this analysis, we report outcomes for sellers who optimize either demand or revenue under a demand-maximizing recommender system. Collectively, these results underscore the influential role of recommender systems in shaping product exposure, influencing seller payoffs, and guiding the evolution of the game.

\section{Conclusions}
\label{sec: conclusion}


AI-based pricing algorithms are widely adopted by sellers due to their minimal requirement for human intervention and their remarkable adaptability to rapidly changing economic conditions. However, both researchers and policymakers have raised concerns regarding the potentially detrimental effects of these algorithms on consumer welfare, particularly in relation to the phenomenon of algorithmic collusion.

In online markets, the high level of digitization facilitates the widespread adoption of pricing algorithms by sellers. Concurrently, platforms extensively employ recommender systems to optimize their own objectives, which can alter product exposure and significantly impact the competitive landscape and pricing dynamics. Despite active research on algorithmic collusion, the role of recommender systems has been largely overlooked. This study aims to address this gap by developing a new repeated game framework that incorporates recommender systems and appropriately adjusts other model components.

After addressing several challenges in specifying and solving the repeated game, we obtain a series of results that hold significant implications for antitrust regulations, platforms, and sellers. Additionally, these findings can also inform and guide future research endeavors in this domain.

For regulators, we demonstrate that the objectives of recommender systems significantly influence price dynamics and equilibrium outcomes, even when AI-based pricing algorithms are present. Given the pivotal role of recommender systems, when antitrust legislation seeks to curb supracompetitive pricing, it may not be sufficient to hold only the AI pricing providers responsible, as current practice often does \citep{robertson2022antitrust, colangelo2021artificial}. It may be beneficial to compare market outcomes both with and without the existing recommendation systems—particularly those that disproportionately favor certain stakeholders (e.g., sellers’ revenue or consumers’ utility in two-sided platforms). Such comparisons would enable regulatory authorities to attribute parts of any supracompetitive pricing not only to the underlying pricing algorithms, but also to the platform’s recommender system. This approach would yield a more equitable and nuanced regulatory framework.

However, it is also important to recognize that a platform’s recommender system often functions as a black box for external regulators. In fact, the pricing algorithms are typically opaque, and the interaction between these algorithms and the recommender system further complicates the analysis. Thus, enhanced AI transparency is a fundamental prerequisite for effective regulation. To promote responsible AI practices and improve regulatory oversight, platforms may need to disclose additional information about the recommender systems they employ, including the specific optimization objectives guiding their design. 

As for the platform, we also find that when a platform’s objective is to enhance consumer utility, increasing the number of products displayed may inadvertently have adverse effects in the presence of pricing algorithms. Moreover, this outcome is moderated by the level of horizontal differentiation among products. Platforms should therefore consider the presence of pricing algorithms and prevailing market conditions (e.g., the degree of horizontal differentiation) when designing recommender systems. For example, platforms like Amazon could tailor their recommendation strategies by category: if a market features highly differentiated products, reducing the number of recommended items might improve long-term consumer satisfaction; in markets with less differentiation, the opposite approach may be more appropriate.

At a broader conceptual level, effective recommender system design in the presence of pricing algorithms may necessitate transitioning from a ``two-tower" recommendation framework \citep{guo2019pal} to a ``three-tower" approach. Traditionally, platforms have focused on understanding consumers’ decision-making processes (one tower) and adjusting the recommendation strategy accordingly (the second tower). However, as pricing algorithms can also respond strategically and dynamically, it is essential to incorporate sellers’ pricing behaviors (the third tower) into the framework. This ensures that recommender systems remain adaptable to changes in both consumer and seller behavior over time.

For sellers that design or adopt pricing algorithms, our findings indicate that the effectiveness of these algorithms is contingent upon the platform's recommendation mechanisms. Consequently, these stakeholders must concurrently evaluate both the pricing algorithms and the platform's revenue-sharing structure, rather than considering them as isolated components. Furthermore, since product differentiation may influence the platform’s recommendation strategies and subsequently affect the rewards garnered by pricing algorithms, it is advantageous for sellers to jointly reconsider both product design and pricing algorithm strategies.

Our study also identifies several future exploration directions for both researchers and practitioners. First, this study is situated within the broader field of the economics of AI, which seeks to evaluate, design, and implement AI systems through the application of economic principles. As AI adoption becomes more widespread, it is increasingly important to consider the economic and societal impacts arising from the interactions among various AI systems. In particular, AI–AI dynamics, such as those between recommender systems and pricing algorithms, should be evaluated jointly rather than in isolation. Our work marks an initial step in examining these interactions and highlights the significance of further investigation into their long-term effects. 

Focusing on our specific domain, we contribute to the algorithmic collusion literature by underscoring the role of recommender systems in shaping algorithmic pricing competition. In doing so, we introduce a new analytical framework that may be valuable for future studies examining similar issues. Besides, our findings also extend the economics of recommender systems literature by incorporating adaptive, AI-based pricing, rather than assuming static or fully informed seller-driven pricing strategies. In other words, the economic impact of recommender systems emerges from their ability to change product exposure and their influence on the dynamic rewards available to pricing algorithms. For empirical researchers and companies with the capacity to conduct large-scale field experiments, examining how price responses evolve (an understudied outcome) under varying recommendation policies would be a valuable future avenue of study. Such efforts also require advancement in causal inference methodologies (including experimental design and suitable evaluation frameworks) to address interference issues arising from dynamic price competition.

Following this line of reasoning, several areas remain for future research. Throughout this study, due to the unavailability of high-frequency pricing data from the sellers' side—consistent with other algorithmic collusion studies across various contexts \citep{dou2024ai, colliard2022algorithmic, calvano2020artificial}—we directly specify the parameters of pricing algorithms. Nevertheless, acquiring detailed pricing data, uncovering sellers’ strategies, empirically quantifying levels of collusion, and conducting additional market design studies represent promising and important avenues for future research.

In addition, we focus on the scenario where consumers are presented with the recommended set and only make their search and purchase decisions among these recommended products (including outside options). In reality, certain consumers may exhibit strong loyalty to specific products and possess extensive knowledge about them. These consumers may purchase these products irrespective of pricing competition or the behavior of recommender systems. Since they do not influence the relative rewards across different pricing decisions of each seller, they do not affect pricing dynamics and can thus be safely excluded from our analyses. There may also be consumers who are already knowledgeable about specific products, potentially encompassing all products within a small market. For these consumers, recommender systems serve solely as tools for additional information discovery. As the proportion of such informed consumers increases, the impact of recommender systems diminishes. In an extreme case where all consumers are aware of all available products in the market beforehand, recommender systems would lose their effectiveness. However, the existence of informed consumers affects only the magnitude, rather than the qualitative direction, of our findings. For the sake of clarity and ease of interpretation, we focus on the setting where consumers view and select from the recommendation set. Future research could explore the heterogeneous welfare impacts of recommender systems and pricing strategies on various types of consumers.

Finally, it is worth noting that translating the economic research of recommender systems into their technical designs necessitates additional efforts \citep{li2024variety, wang2024recommending, bi2024consumer, li2023hierarchical}. Although this endeavor is beyond the scope of the present study, we encourage algorithm designers in both academia and industry to recognize and account for the strategic pricing decisions enabled by AI-driven pricing algorithms. By doing so, they can develop algorithms that enhance both the overall welfare and the equitable distribution of benefits among stakeholders. 

\newpage
\bibliography{thebibliography}

\bibliographystyle{informs2014}

%

\newpage
\begin{APPENDICES}

\renewcommand{\thesection}{\Alph{section}}
\renewcommand{\thesubsection}{\Alph{section}.\arabic{subsection}}
\counterwithin{table}{section}
\counterwithin{figure}{section}
\renewcommand{\thetable}{\Alph{section}\arabic{table}}
\renewcommand{\thefigure}{\Alph{section}\arabic{figure}}

\newpage
\section{Consumer Search and Purchase Process}
\label{appendix:visualization}

\begin{figure}[ht]
  \centering
  \includegraphics[width=1\textwidth]{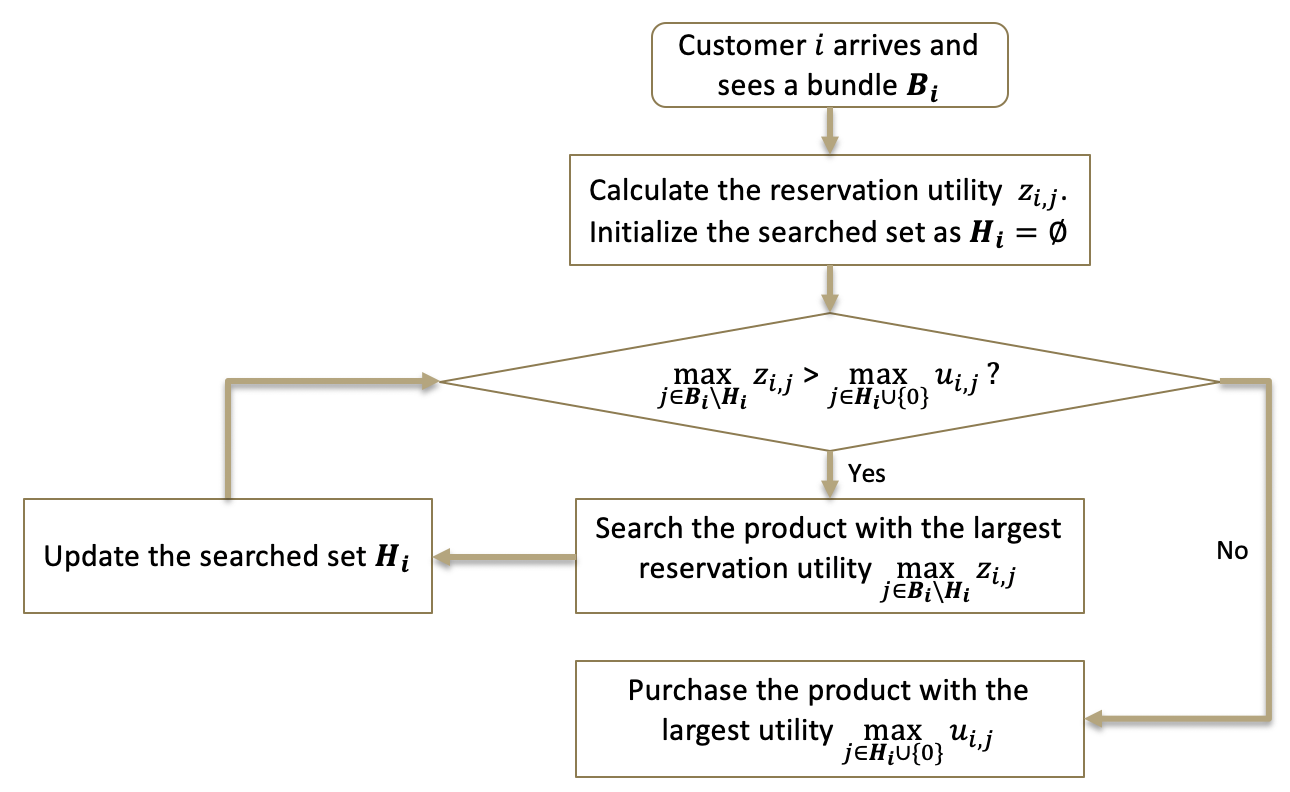}
  \label{fig:visualization}
  \caption{Consumer Search and Purchase Process}
\end{figure}

\newpage
\section{Alternative Economic Environment: Logit Demand}
\label{appendix:logit}

To align our proposed structural search model with the traditional algorithmic collusion literature \citep{calvano2020artificial}, which typically employs the canonical logit model, we also evaluate our results using the logit model as a robustness check. In fact, the logit demand model can be considered a restricted version derived from the sequential search model. Specifically, from the sequential search model, we can (1) set the search cost to zero instead of allowing it to take any value estimated from data, (2) assume that all recommended products' full information is available to consumers, thereby eliminating position bias under this full information perspective (a natural consequence of (1)), and (3) change the distribution of taste shocks from a normal distribution to a type-I extreme value distribution to facilitate a closed-form expression of the choice probability as a logit function. This modification yields the logit choice model when a consumer $i$ sees a recommendation set:
\begin{equation}
    d_{i,j}^t = \frac{e^\frac{{v_j - p_j^t}}{\mu}}{{\sum_{k \in \bm{B_i}} e^\frac{{v_k - p_k^t}}{\mu}  + e^\frac{v_{0}}{\mu}}}
\end{equation}

Here, the parameters $v_j$ represent representative utility for each product $j$, capturing vertical differentiation. The parameter \(\mu\) reflects the degree of horizontal differentiation where perfect substitution is achieved as \(\mu\) approaches $0$. The denominator of the equation represents the exponential sum over all products' consumption utilities within the recommendation set $B_i$ along with the outside good, characterized by the quality index $v_0$. Furthermore, under the assumption that all consumers are homogeneous in their observable preferences and differ solely in their idiosyncratic taste shocks, the above single-individual choice probability also characterizes the demand for product \( j \) when a unit of consumers is presented with the recommendation set \( \bm{B}_i \).

In alignment with our primary framework, we continue to address a two-out-of-three recommendation scenario. In the absence of position bias within this logit demand model, product ranking does not influence consumer behavior, thereby reducing the platform's task to an assortment optimization problem. Furthermore, since consumers are perceived as homogeneous by the platform, it is sufficient to construct a single recommendation set that either maximizes revenue or maximizes consumer utility to all consumers.

\begin{table}[h]
\centering
\caption{Price, Revenue, and Utility Comparison (Logit Demand Model)}
\label{tab:logit}
\newcolumntype{L}[1]{>{\raggedright\arraybackslash}p{#1}}
\newcolumntype{C}[1]{>{\centering\arraybackslash}p{#1}}
\newcolumntype{R}[1]{>{\raggedleft\arraybackslash}p{#1}}
\renewcommand\arraystretch{1.5} 
\begin{tabular}{L{1.5cm}C{4cm}C{4cm}C{4cm}}
\hline 
\hline
 \textbf{ } & \textbf{Revenue-Maximization} & \textbf{Utility-Maximization} & \textbf{Baseline} 
\\
 \hline
 Prices & 0.8829 (0.0051) & 0.3071 (0.0194) & 0.5553 (0.0128) \\
 Revenue & 0.2243 (0.0002) & 0.0703 (0.0046) & 0.1630 (0.0033) \\
 Utility & 0.3543 (0.0012) & 0.9525 (0.0150) & 0.6499 (0.0117) \\
\hline \hline
\end{tabular}
\end{table}

The expected revenue for each recommendation set can be computed as:
\begin{equation}
    r_{\bm{B}_i} = \sum_{j \in \bm{B}_i} p_j^t \cdot \frac{\exp\left(\frac{v_j - p_j^t}{\mu}\right)}{\sum_{k \in \bm{B}_i} \exp\left(\frac{v_k - p_k^t}{\mu}\right) + \exp\left(\frac{v_{0}}{\mu}\right)}
\end{equation}

Conversely, the expected utility for a recommendation set is calculated using the welfare formula \citep{train2009discrete}:
\begin{equation}
    TU_{\bm{B}_i} = \sum_{k \in \bm{B}_i} \exp\left(\frac{v_k - p_k^t}{\mu}\right) + \exp\left(\frac{v_{0}}{\mu}\right)
\end{equation}

We adhere to the game sequence outlined in Algorithm \ref{alg1} of Section \ref{subsec: framework} and employ Q-learning pricing algorithms with consistent parameters. We normalize the valuations \( v_j \) of all products to 1 and set the outside option valuation \( v_0 \) to 0, allowing prices to fluctuate between 0 and 1. After conducting 50 simulation trials, each spanning 3,000,000 periods, we present the equilibrium prices, revenue, and utility in a manner consistent with Section \ref{sec: results}.

As illustrated in Table \ref{tab:logit}, the comparisons across the three cases are qualitatively similar to our main results, thereby reinforcing the robustness of our findings across different demand model specifications.

\newpage
\section{Alternative Economic Environment: Heterogeneous Logit Demand}
\label{appendix:heterologit}

In Appendix \ref{appendix:logit}, we employ a logit demand model that does not incorporate customer segmentation \citep{calvano2020artificial}. In this appendix, we extend the economic environment by introducing heterogeneous customer types and subsequently adapt the recommender systems to optimize for different customer categories.

Specifically, we allow each consumer to possess heterogeneous outside option values. To illustrate, we assign half of the consumers a higher outside option value of \( v_0 = 0.25 \), while the remaining consumers have a lower outside option value of \( v_0 = 0 \). For each segment, we recommend the set that maximizes either expected revenue or expected utility. With all other settings held constant, we present the results in Table \ref{tab:logit_hetero}. 

Notably, the primary findings remain consistent under this modified setting. When the recommender system is configured for revenue maximization, sellers tend to engage in more collusive behavior, resulting in elevated equilibrium prices (0.7979) and increased revenue (0.1907) at the expense of consumer utility (0.4563). Conversely, when the system prioritizes the maximization of consumer utility, the equilibrium is characterized by lower prices (0.5229) and enhanced consumer utility (0.6925), alongside a reduction in seller revenue (0.1469).

\begin{table}[ht]
\centering
\caption{Price, Revenue, and Utility Comparison (Heterogeneous Logit Demand Model)}
\label{tab:logit_hetero}
\newcolumntype{L}[1]{>{\raggedright\arraybackslash}p{#1}}
\newcolumntype{C}[1]{>{\centering\arraybackslash}p{#1}}
\newcolumntype{R}[1]{>{\raggedleft\arraybackslash}p{#1}}
\renewcommand\arraystretch{1.5} 
\begin{tabular}{L{1.5cm}C{4cm}C{4cm}C{4cm}}
\hline 
\hline
 \textbf{ } & \textbf{Revenue-Maximization} & \textbf{Utility-Maximization} & \textbf{Baseline} 
\\
 \hline
 Prices & 0.7979 (0.0086) & 0.3075 (0.0189) & 0.5229 (0.0105) \\
 Revenue & 0.1907 (0.0004) & 0.0659 (0.0039) & 0.1469 (0.0024) \\
 Utility & 0.4563 (0.0050) & 0.9666 (0.0131) & 0.6925 (0.0091) \\
\hline \hline
\end{tabular}
\end{table}

\newpage
\section{Empirical Data Description and Summary Statistics}
\label{appendix:data}

In this appendix, we detail the construction of our dataset for structural estimation, basically following the steps taken in \cite{chung2024simulated} and \cite{ ursu2018power}. The original Expedia dataset contains millions of search impressions, including query information and all displayed hotels. Because the dataset is divided into training and test sets, and the test set lacks consumer decision data (search or purchase), we focus exclusively on the training set.

We apply the following filtering procedures to the training set:

\begin{enumerate}
    \item Exclude search impressions that include any hotels with extreme prices (less than \$10 or more than \$1,000).
    \item For search impressions that led to transactions, infer the tax from the total spending and displayed prices, and exclude those impressions where the inferred tax is less than \$1 or exceeds 30\% of the price.
    \item Drop search impressions containing any missing values (NaN) in query or hotel attributes.
    \item Exclude search impressions based on Expedia's ranking rather than the random ranking.
    \item Segment the dataset by travel destination, recognizing that consumers and hotels may vary significantly across destinations, and focus on the destination with the highest volume of search impressions (destination ID 8192).
    \item Although most impressions have similar lengths (number of hotels displayed), there are outliers. For the selected destination, the number of hotels per impression ranges from 5 to 35, with the 25th percentile at 29. Therefore, we restrict our analysis to search impressions with lengths greater than 29\footnote{This is slightly less stringent than \cite{ursu2018power}, which only uses the two most frequently occurring search impression lengths.}.
\end{enumerate}

The final dataset comprises 1,067 impressions and 34,702 observations, where each observation represents a hotel option within a search impression. It includes hotel information, query information (customer attributes constant across different hotels within one search impression), and final decisions (search or purchase). Additionally, we add an outside option for each search impression, coded as a purchase decision equal to one if the customer does not buy from any displayed hotel and zero otherwise. Thus, for the structural estimation, the total number of observations is 35,769 (34,702 hotel options plus 1,067 outside options). For 34,702 hotel options, the variable descriptions and summary statistics are provided in Table \ref{tab:data}.

\newpage
\begin{table}[ht]
\centering
\caption{Variable Description and Summary Statistics}
\label{tab:data}
\newcolumntype{L}[1]{>{\raggedright\arraybackslash}p{#1}}
\newcolumntype{C}[1]{>{\centering\arraybackslash}p{#1}}
\newcolumntype{R}[1]{>{\raggedleft\arraybackslash}p{#1}}
\renewcommand\arraystretch{1.5} 
\begin{tabular}{L{3.5cm}L{6cm}C{1.2cm}C{1.2cm}C{1.2cm}C{1.2cm}}
\hline 
\hline
 \textbf{Variable} & \textbf{Description} & \textbf{Mean} & \textbf{Std} & \textbf{Min} & \textbf{Max}
\\
 \hline
 Price (\$100) & Price of the hotel (\$100) & 1.4326 & 0.9026 & 0.1700 & 9.8400\\
 HotelStars & The star rating of the hotel & 3.9291 & 0.8633 & 2 & 5\\
 HotelReview & Average customer review rating & 4.0370 & 0.4901 & 3 & 5\\
 Brand & A binary indicator of chained hotel & 0.7872 & 0.4093 & 0 & 1\\
 HotelLocationScore & A score of location desirability & 4.0328 & 0.3011 & 3.0400 & 4.3800\\
 Promo & A binary indicator of promotion & 0.5919 & 0.4915 & 0 & 1\\
 ReqRoomNum & Number of rooms required & 1.1414 & 0.4151 & 1 & 4\\
 SatNight & A binary indicator of Saturday night involved in travel & 0.4752 & 0.4994 & 0 & 1\\
 Click & A binary indicator of whether the customer search the hotel& 0.0370 & 0.1888 & 0 & 1\\
 Tran & A binary indicator of whether the customer book the hotel& 0.0026 & 0.0511 & 0 & 1\\
\hline \hline
\end{tabular}
\end{table}

\newpage
\section{Alternative Pricing Algorithm: More Exploration}
\label{appendix:explore_more}

In our primary analyses, we demonstrate that prices increase (decrease) under conditions where product exposure is governed by revenue-maximizing (utility-maximizing) recommender systems. As illustrated in Figure \ref{fig:learn_base}, the algorithms' exploration process appears to alter only the magnitude of price changes rather than their direction across the three different scenarios examined. Nevertheless, it remains plausible that extended and more comprehensive exploration could yield different outcomes, such as the algorithms learning to collude even in the presence of utility-maximizing recommender systems. Therefore, in this appendix, we investigate the effects of increased exploration on price dynamics and equilibrium outcomes.

Using an $\epsilon$-greedy algorithm, the exploration rate, denoted by $\epsilon$, modulates the probability of undertaking a stochastic action at each temporal interval according to the equation $\epsilon = e^{-\omega t}$. This probability undergoes exponential decay, governed by both the decay rate $\omega$ and the time parameter $t$. Extended exploration periods enable the algorithm to obtain more accurate estimates of the optimal pricing strategy; however, they also incur higher opportunity costs due to the exploration of suboptimal pricing alternatives. In our baseline configuration, the decay rate $\omega$ is set to $1.5 \times 10^{-6}$. In contrast, this appendix explores a ``More Exploration" scenario, wherein we employ a substantially smaller decay rate of $\omega = 5 \times 10^{-8}$, thereby significantly increasing exploratory activities. For example, after three million periods, the pricing algorithms in the baseline configuration retain an approximate probability of $1.11\%$ for opting for a stochastic action. In contrast, in the ``More Exploration" scenario, this probability remains above $80\%$.

To ensure experimental consistency, all parameters remain identical to those in the baseline configuration, with the sole modification being the exploration rate parameter $\omega$. We conduct 20 simulation runs, each spanning 80,000,000 periods, after which the exploration probability falls below $1\%$. The resulting average equilibrium prices and profits are presented in Table \ref{tab:table_sufficient_mu025}, while the temporal evolution of the learning dynamics is depicted in Figure \ref{fig:learn_sufficient_mu025}. Notably, the equilibrium prices attained in the ``More Exploration" scenario exhibit substantial congruence with those observed in the baseline scenario. Specifically, compared to the equilibrium price of 1.6861 observed in the baseline condition, the implementation of revenue-maximizing recommender systems results in an elevated price of 2.3734. Conversely, the deployment of utility-maximizing recommender systems leads to a reduced price of 1.6861. Furthermore, the learning dynamics of the pricing algorithms, influenced by the different types of recommender systems, continue to progress monotonically toward their respective equilibria, in alignment with the objectives of the respective recommender systems.

\newpage
\begin{table}[ht]
\centering
\caption{Price, Revenue, and Utility Comparison ($\mu = 0.25$, $\omega = 5 \times 10^{-8}$)}
\label{tab:table_sufficient_mu025}
\newcolumntype{L}[1]{>{\raggedright\arraybackslash}p{#1}}
\newcolumntype{C}[1]{>{\centering\arraybackslash}p{#1}}
\newcolumntype{R}[1]{>{\raggedleft\arraybackslash}p{#1}}
\renewcommand\arraystretch{1.5} 
\begin{tabular}{L{1.5cm}C{4cm}C{4cm}C{4cm}}
\hline 
\hline
 \textbf{ } & \textbf{Revenue-Maximization} & \textbf{Utility-Maximization} & \textbf{Baseline} 
\\
 \hline
 Prices & 2.3734 (0.0225) & 0.8260 (0.0665) & 1.6861 (0.0463) \\
 Revenue & 0.2578 (0.0006) & 0.1067 (0.0052) & 0.2165 (0.0028) \\
 Utility & 0.3746 (0.0017) & 0.6664 (0.0078) & 0.4740 (0.0062) \\
\hline \hline
\end{tabular}
\end{table}
    
\begin{figure}[H]
\centering
\subfloat[Revenue-Maximization RS]{\includegraphics[width=.5\linewidth]{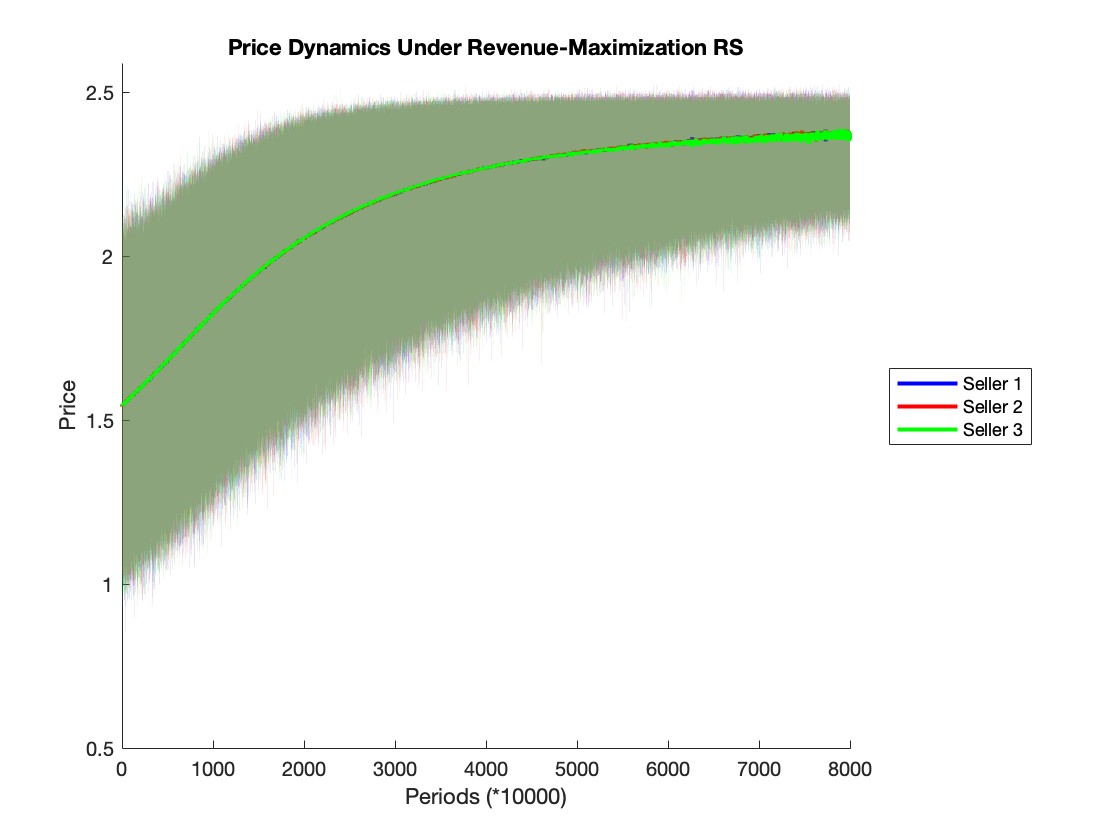}}\hfill
\subfloat[Utility-Maximization RS]{\includegraphics[width=.5\linewidth]{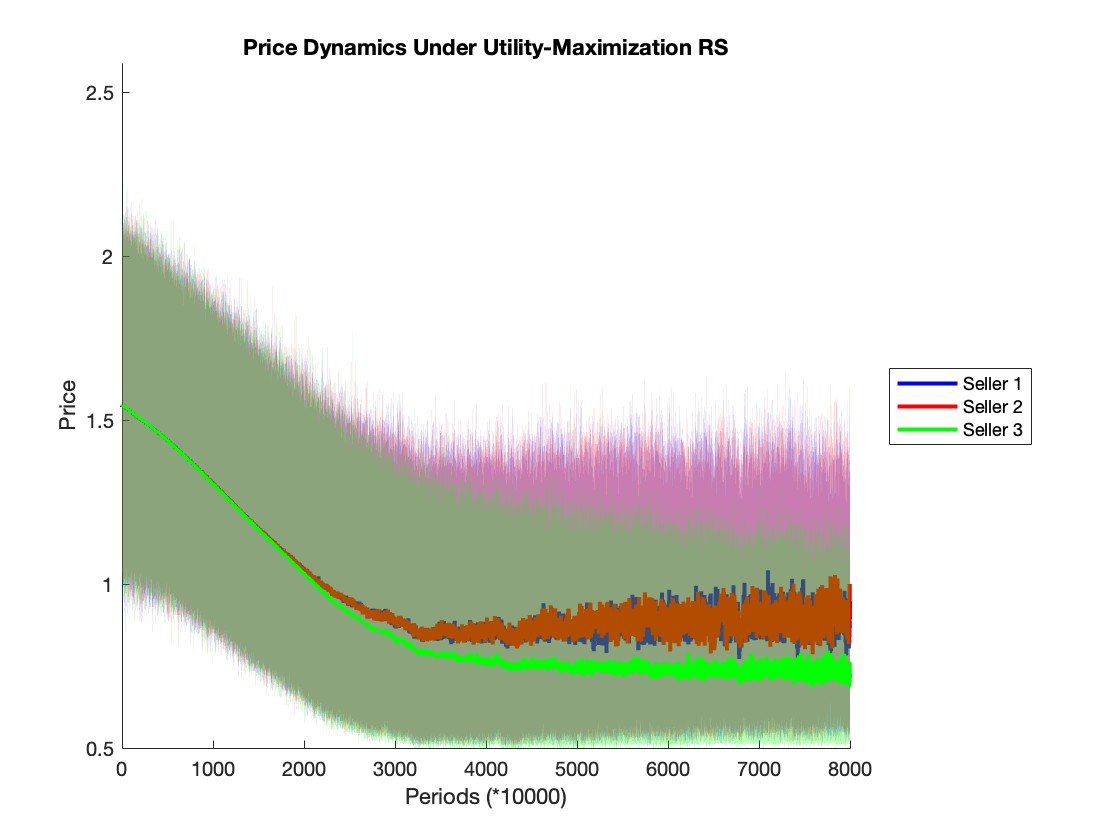}}\par 
\subfloat[Baseline]{\includegraphics[width=.5\linewidth]{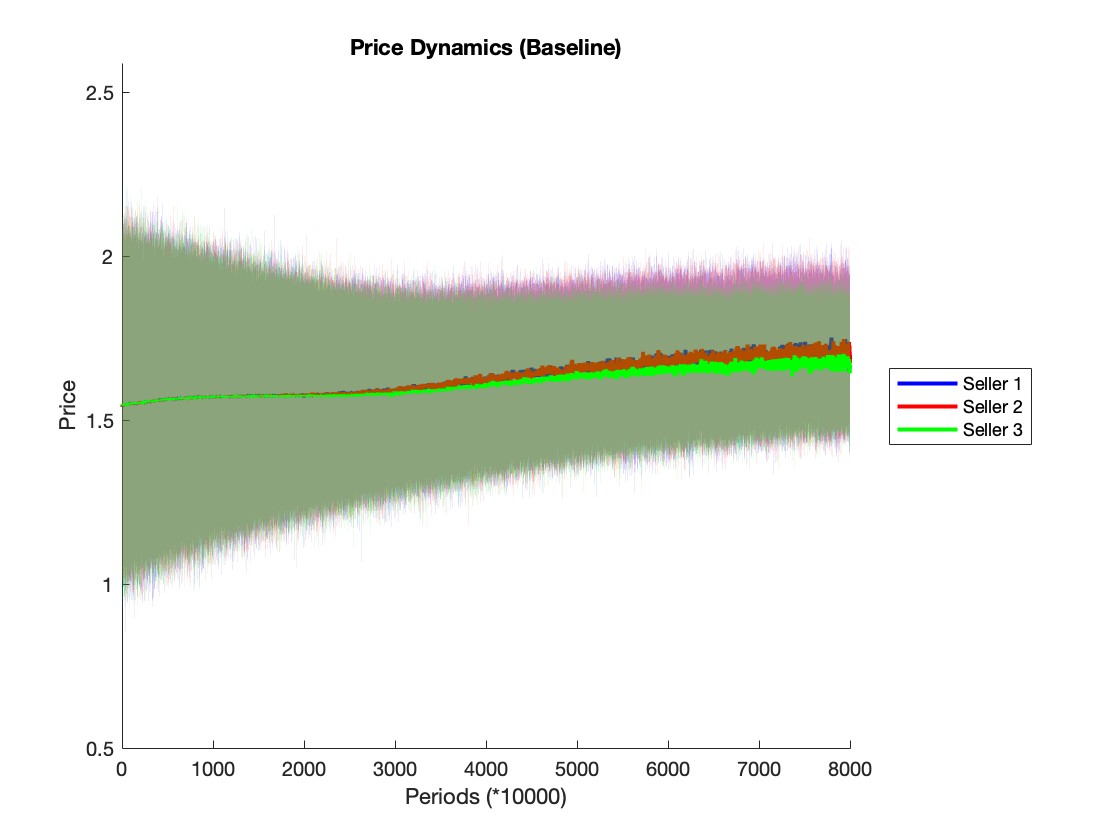}}\\
\caption{Learning Dynamics  ($\mu = 0.25$, $\omega = 5 \times 10^{-8}$)}
\medskip
\begin{minipage}{0.8\textwidth}
{\footnotesize  Note: The points on the continuous lines depict the moving average, with a window size of 10000, of the average price observed across 20 experiments at each $t$. The shaded region's upper and lower limits indicate the highest and lowest values occurring within the window.\par}
\end{minipage}
\label{fig:learn_sufficient_mu025}
\end{figure}

\newpage
\section{Alternative Pricing Algorithm: Multi-armed Bandit}
\label{appendix:mab}

Q-learning inherently incorporates a forward-looking component through its state transitions. To assess the robustness of our findings, we relax the forward-looking assumption by modeling the algorithms as basic multi-armed bandits. Instead of constructing a Q-table of size \( |\bm{S}| \times |\bm{P}| \) to record the rewards of various pricing decisions under different states (i.e., past pricing outcomes), we now construct a reward table of size \( 2 \times |\bm{P}| \), which tracks the number of times each pricing decision (arm) has been attempted and the corresponding average reward for each arm. We maintain an initially empty reward table for each pricing agent (seller) and allow each seller to learn to set prices through sampling. Consistent with our Q-learning approach, we employ an $\epsilon$-greedy sampling strategy with the same exploration rate.

We follow the simultaneous game sequence once more and present the equilibrium results in Table \ref{tab:table_bandits} and the price dynamics in Figure \ref{fig:learn_bandits}. Notably, the results remain largely consistent with our primary findings. This consistency arises because our core mechanism relies on the recommender system's influence on the pricing agents' rewards, making forward-looking behavior an unnecessary condition for the pricing algorithms to produce the observed outcomes.

\begin{table}[ht]
\centering
\caption{Price, Revenue, and Utility Comparison Across Objectives (MAB)}
\label{tab:table_bandits}
\newcolumntype{L}[1]{>{\raggedright\arraybackslash}p{#1}}
\newcolumntype{C}[1]{>{\centering\arraybackslash}p{#1}}
\newcolumntype{R}[1]{>{\raggedleft\arraybackslash}p{#1}}
\renewcommand\arraystretch{1.5} 
\begin{tabular}{L{1.5cm}C{4cm}C{4cm}C{4cm}}
\hline 
\hline
 \textbf{ } & \textbf{Revenue-Maximization} & \textbf{Utility-Maximization} & \textbf{Baseline} 
\\
 \hline
 Prices & 2.3393 (0.0149) & 0.9881 (0.0118) & 1.6596 (0.0057) \\
 Revenue & 0.2576 (0.0003) & 0.1544 (0.0013) & 0.2195 (0.0007) \\
 Utility & 0.3805 (0.0015) & 0.6118 (0.0018) & 0.4753 (0.0010) \\
\hline \hline
\end{tabular}
\end{table}

\newpage
\begin{figure}[h]
\centering
\subfloat[Revenue-Maximization RS]{\includegraphics[width=.5\linewidth]{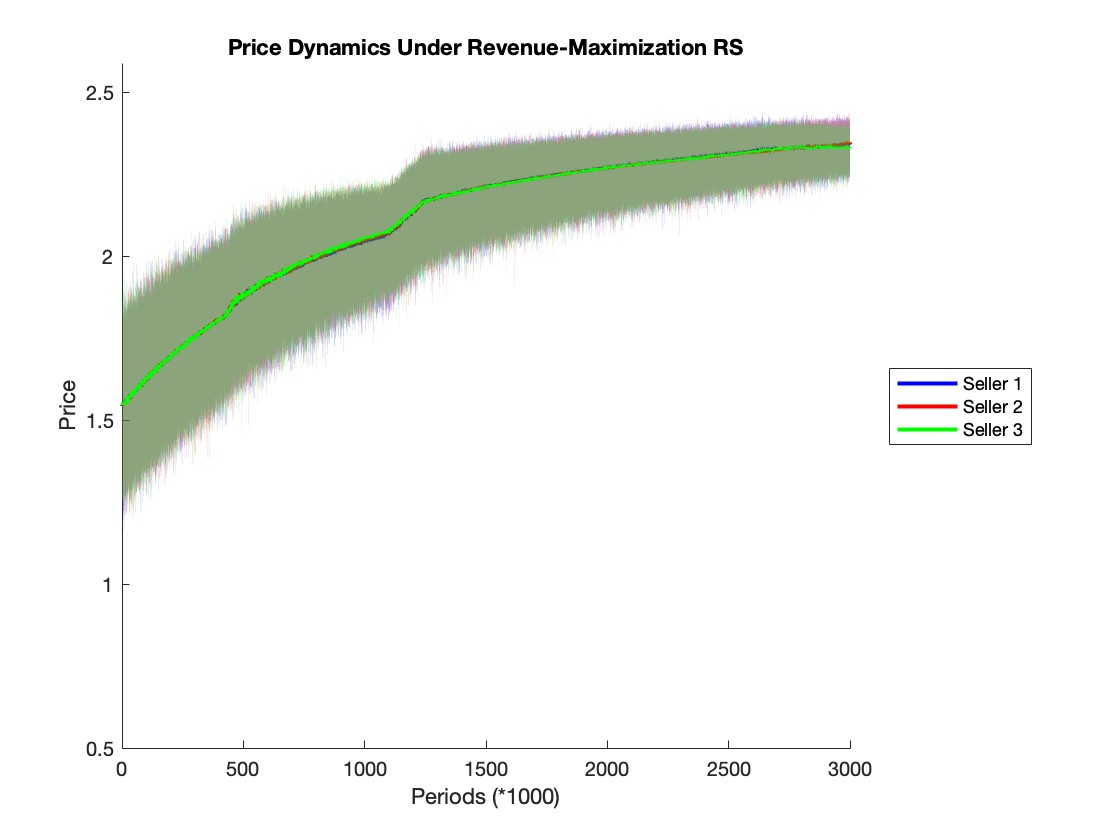}}\hfill
\subfloat[Utility-Maximization RS]{\includegraphics[width=.5\linewidth]{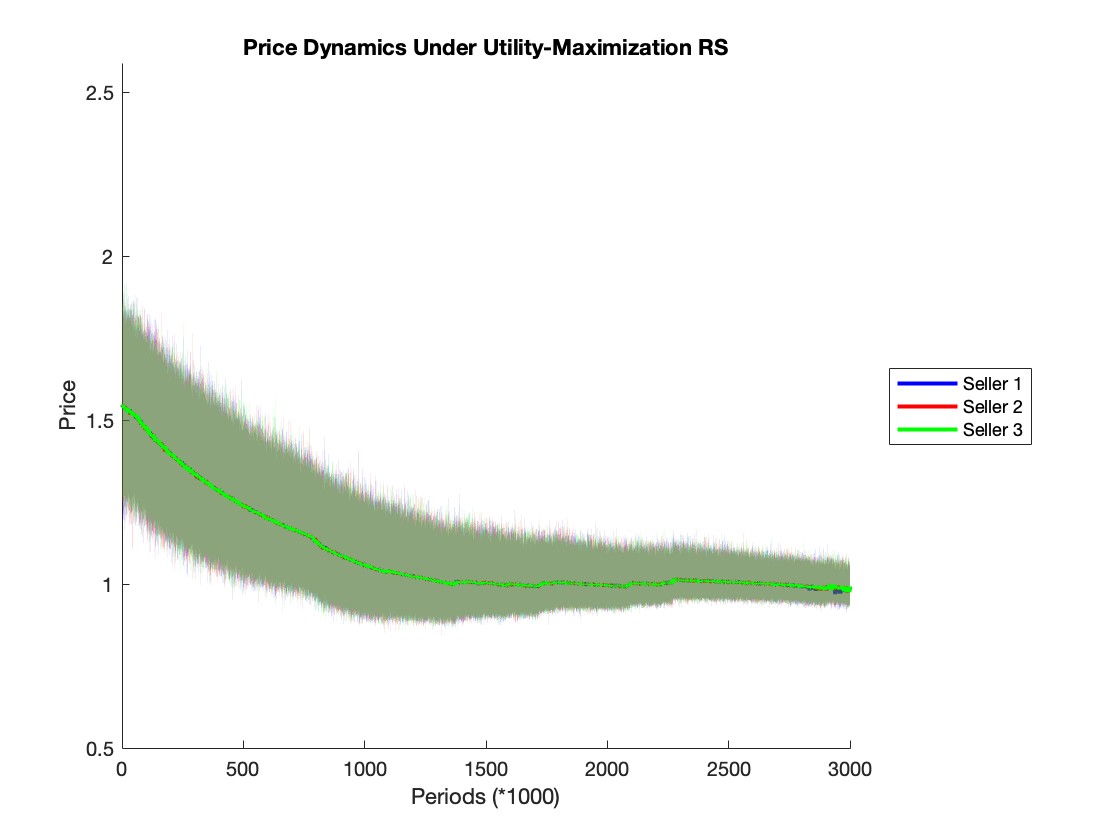}}\par 
\subfloat[Baseline]{\includegraphics[width=.5\linewidth]{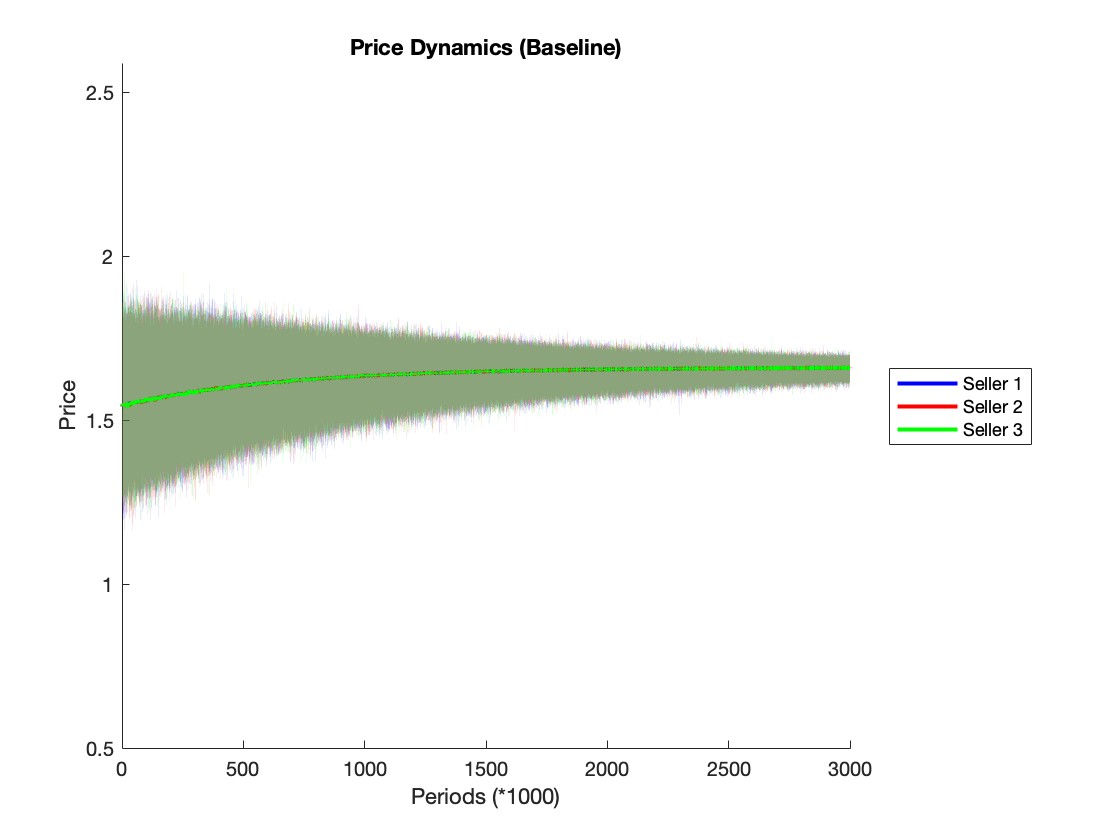}}\\
\caption{Learning Dynamics (MAB)}
\medskip
\begin{minipage}{0.8\textwidth}
{\footnotesize  Note: The points on the continuous lines depict the moving average, with a window size of 1000, of the average price observed across 50 experiments at each $t$. The shaded region's upper and lower limits indicate the highest and lowest values occurring within the window.\par}
\end{minipage}
\label{fig:learn_bandits}
\end{figure}

\newpage
\section{Alternative Pricing Objective of Sellers}
\label{appendix:pricingobj}

In \cite{calvano2020artificial}, sellers are assumed to optimize revenue. When sellers maximize revenue, they must balance a trade-off between demand and per-sale revenue: setting lower prices can boost demand but reduce revenue per sale, whereas setting higher prices can increase revenue per sale but suppress demand. In such contexts, sellers may engage in price undercutting, thereby intensifying price competition.

Accordingly, starting from the Bertrand game, most studies in industrial organization and traditional game theory assume that sellers aim to maximize revenue, ensuring meaningful competitive dynamics \citep{narahari2009game}. By contrast, if a seller instead sought to maximize demand (or consumer utility), it could simply set its price to zero, regardless of competitors’ prices, thereby removing any incentive for strategic interactions.

Despite these considerations, we show that recommender systems retain their influence even when sellers optimize demand and willingly offer the lowest possible prices. To illustrate this, we modify the seller’s objective to demand maximization and rerun our numerical experiments. As reported in Table \ref{tab:table_optdemand}, both the utility-maximizing and the baseline scenarios produce the lowest possible prices. Nevertheless, when the platform pursues revenue maximization, it still induces sellers to sustain relatively higher prices. In fact, these prices even exceed the equilibrium levels observed under baseline conditions with revenue-driven sellers. This finding further emphasizes the pivotal role of recommender systems in shaping algorithmic pricing strategies.

Furthermore, when some sellers prioritize demand while others focus on revenue maximization, an asymmetric equilibrium might emerge. For example, consider a scenario in which one seller aims to maximize demand and two sellers aim to maximize revenue, with the recommender system oriented towards utility maximization. In this case, the utility-driven seller reduces its price to the lowest possible level, while the revenue-driven sellers converge to slightly higher prices, though these prices remain lower than those observed in the baseline scenario. Due to the vast number of possible combinations, we do not present all these results here. However, from all these interesting asymmetric equilibria, we consistently observe the influential role of recommender systems across various configurations.

\begin{table}[ht]
\centering
\caption{Price, Revenue, and Utility Comparison Across Objectives (Demand-Driven Sellers)}
\label{tab:table_demandseller}
\newcolumntype{L}[1]{>{\raggedright\arraybackslash}p{#1}}
\newcolumntype{C}[1]{>{\centering\arraybackslash}p{#1}}
\newcolumntype{R}[1]{>{\raggedleft\arraybackslash}p{#1}}
\renewcommand\arraystretch{1.5} 
\begin{tabular}{L{1.5cm}C{4cm}C{4cm}C{4cm}}
\hline 
\hline
 \textbf{ } & \textbf{Revenue-Maximization} & \textbf{Utility-Maximization} & \textbf{Baseline} 
\\
 \hline
 Prices & 1.8128 (0.0062) & 0.5109 (0.0103) & 0.5109 (0.0103) \\
 Revenue & 0.2389 (0.0002) & 0.0901 (0.0001) & 0.0914 (0.0005) \\
 Utility & 0.4449 (0.0005) & 0.6974 (0.0004) & 0.6654 (0.0032) \\
\hline \hline
\end{tabular}
\end{table}

\newpage
\section{Alternative Game Sequence: Asynchronous Pricing Game}
\label{appendix:asyn}

In our primary analyses, we adhere to the framework established by \citet{calvano2020artificial}, assuming that all sellers make pricing decisions simultaneously. While this assumption underpins our main analyses, we revisit and relax it in this section.

If we permit sellers to update their pricing decisions sequentially—following the order of seller 1, seller 2, seller 3, and so forth—the decision-making process unfolds as follows: seller 1 updates, followed by seller 2 and then seller 3. Considering the structure of Q-learning, seller 1 observes the prices from the preceding three periods before making a decision. Afterwards, the environment changes in the subsequent two periods following seller 1’s decision. By aggregating these three periods into a single ``macro'' period, the scenario reverts to the simultaneous decision-making framework. Therefore, from a theoretical standpoint, an asynchronous game sequence does not significantly alter the nature of the algorithmic pricing game.

This reasoning similarly applies to settings where multi-armed bandits (MAB) serve as the pricing algorithms. For the sake of comparison, we present the results of the MAB approach under the aforementioned asynchronous game in Table \ref{tab:table_bandits_asyn}. As anticipated, these results are largely consistent with those reported in Appendix \ref{appendix:mab}, thereby reinforcing the robustness of our findings. The same consistency is observed in the Q-learning based pricing game.

\begin{table}[ht]
\centering
\caption{Price, Revenue, and Utility Comparison Across Objectives (Asynchronous)}
\label{tab:table_bandits_asyn}
\newcolumntype{L}[1]{>{\raggedright\arraybackslash}p{#1}}
\newcolumntype{C}[1]{>{\centering\arraybackslash}p{#1}}
\newcolumntype{R}[1]{>{\raggedleft\arraybackslash}p{#1}}
\renewcommand\arraystretch{1.5} 
\begin{tabular}{L{1.5cm}C{4cm}C{4cm}C{4cm}}
\hline 
\hline
 \textbf{ } & \textbf{Revenue-Maximization} & \textbf{Utility-Maximization} & \textbf{Baseline} 
\\
 \hline
 Prices & 2.3566 (0.0147) & 0.9677 (0.0081) & 1.6598 (0.0062) \\
 Revenue & 0.2578 (0.0003) & 0.1513 (0.0004) & 0.2195 (0.0007) \\
 Utility & 0.3783 (0.0015) & 0.6162 (0.0005) & 0.4753 (0.0010) \\
\hline \hline
\end{tabular}
\end{table}


\newpage
\section{Alternative Recommendation Objective: Total Demand}
\label{appendix:optdemand}

It is worth noting that demand can serve as another plausible objective for the platform. Economically, demand is analogous to utility, since higher utility increases consumers’ purchase probability and, consequently, overall demand. To maintain conceptual clarity and alignment with the existing literature, we primarily focus on revenue and consumer utility in our main analyses, while nonetheless examining demand as an alternative objective here. Specifically, we substitute the platform’s objective with total market demand, keeping all other aspects of the recommender system unchanged.

We conduct two sets of numerical experiments, one with revenue-driven sellers and another with demand-driven sellers, both operating under a demand-maximizing recommender system. As reported in Table \ref{tab:table_optdemand}, our rationale holds. Under the standard revenue-driven pricing framework, the equilibrium price settles at a level lower than the baseline scenario (see Table \ref{tab:table_base} for comparison). When sellers optimize demand, as expected, the price converges to its minimum possible level, consistent with the outcomes shown in Table \ref{tab:table_demandseller}.

\begin{table}[ht]
\centering
\caption{Price, Revenue, and Utility Comparison Across Different Pricing Objectives}
\label{tab:table_optdemand}
\newcolumntype{L}[1]{>{\raggedright\arraybackslash}p{#1}}
\newcolumntype{C}[1]{>{\centering\arraybackslash}p{#1}}
\newcolumntype{R}[1]{>{\raggedleft\arraybackslash}p{#1}}
\renewcommand\arraystretch{1.5} 
\begin{tabular}{L{1.5cm}C{5cm}C{5cm}}
\hline 
\hline
 \textbf{} & \textbf{Revenue-Driven Sellers} & \textbf{Demand-Driven Sellers}
\\
 \hline
 Prices & 1.0992 (0.0101) & 0.5109 (0.0103)\\
 Revenue & 0.1726 (0.0008) & 0.0930 (0.0007)\\
 Utility & 0.5633 (0.0013) & 0.6717 (0.0002)\\
\hline \hline
\end{tabular}
\end{table}

\newpage
\section{Alternative Recommendation Objective: Weighted Average}
\label{appendix:optweight}

It is worth noting that, in practice, platforms may adopt a mixed objective to balance the welfare of both sellers (revenue) and consumers (utility). To illustrate how varying the weight assigned to these two objectives influences the pricing dynamics, we examine five different cases with weights ranging from 0 (pure utility maximization) to 1 (pure revenue maximization) and conduct corresponding numerical experiments. Our findings indicate that as the weight increases, equilibrium prices and sellers' revenues also increase, while consumer utility gradually decreases. Given the monotonic relationship between the weights and equilibrium prices, we infer that when the weight is set between 0.25 and 0.5, the equilibrium price aligns with the baseline scenario (with no optimization, as shown in Table \ref{tab:table_base}). This result further underscores the pivotal role of platform recommendations in shaping market outcomes.

\begin{table}[ht]
\centering
\caption{Price, Revenue, and Utility Comparison Across Objectives (Different Weights)}
\label{tab:table_weighted}
\newcolumntype{L}[1]{>{\raggedright\arraybackslash}p{#1}}
\newcolumntype{C}[1]{>{\centering\arraybackslash}p{#1}}
\newcolumntype{R}[1]{>{\raggedleft\arraybackslash}p{#1}}
\renewcommand\arraystretch{1.5} 
\begin{tabular}{L{1.5cm}C{2.5cm}C{2.5cm}C{2.5cm}C{2.5cm}C{2.5cm}}
\hline 
\hline
 \textbf{Weight} & \textbf{0} & \textbf{0.25} & \textbf{0.5} & \textbf{0.75} & \textbf{1} 
\\
 \hline
 Prices & 0.9095 (0.0478) & 1.4257 (0.0062) & 2.1463 (0.0168) & 2.2925 (0.0188) & 2.3844 (0.0137)\\
 Revenue & 0.1174 (0.0041) & 0.2079 (0.0009) & 0.2468 (0.0004) & 0.2566 (0.0004) & 0.2583 (0.0003)\\
 Utility & 0.6506 (0.0063) & 0.5388 (0.0010) & 0.4284 (0.0014) & 0.3940 (0.0016) & 0.3749 (0.0012)\\
\hline \hline
\end{tabular}
\end{table}

\end{APPENDICES}

%
%







\end{document}